%% file: main_aaai.tex
\documentclass[letterpaper]{article} 
\usepackage{aaai2026}  
\usepackage{times}  
\usepackage{helvet}  
\usepackage{courier}  
\usepackage[hyphens]{url}  
\usepackage{graphicx} 
\usepackage{natbib}  
\usepackage{caption} 
\usepackage{algorithm}
\usepackage{algorithmic}

\usepackage{newfloat}
\usepackage{listings}

\DeclareCaptionStyle{ruled}{labelfont=normalfont,labelsep=colon,strut=off} 
\floatstyle{ruled}
\newfloat{listing}{tb}{lst}{}
\floatname{listing}{Listing}
\input{config}

\title{DynamicRTL: RTL Representation Learning for Dynamic Circuit Behavior}

\input{sections/0-author}

\usepackage{bibentry}

\begin{document}

\maketitle

\input{sections/0-abstract}

\input{sections/1-introduction}

\input{sections/2-background}


\input{sections/4-model}

\input{sections/5-experiment}
\input{sections/5.1-downstream-exp}

\input{sections/7-conclusion}


\newpage

\section*{Acknowledgement}
This work was partly supported by 
the National Natural Science Foundation of China (Grant No. 62090021), 
the National Key Laboratory for Multimedia Information Processing (Peking University), 
the Hong Kong Research Grants Council (RGC) under Grant No. 14212422, 14202824, and C6003-24Y, 
and in part by Huawei Technolgies Co. Ltd. under grant No. N2-2c-TH2420350.

\bibliography{aaai2026}

\newpage
\setcounter{secnumdepth}{2}

\input{appendix/appendix}

\end{document}

%% file: config.tex
\usepackage{graphicx}
\usepackage{amssymb}
\usepackage{booktabs}
\usepackage{caption}
\usepackage{array}
\usepackage{subcaption}

\usepackage{tabularx}
\usepackage{threeparttable}

\usepackage{multirow}
\usepackage{placeins}

\def\eg{\textit{e.g.,}\ }
\def\ie{\textit{i.e.,}\ }

\newcommand{\cmd}[1]{\texttt{\small{#1}}}

\usepackage{microtype}
\usepackage{booktabs} 

\usepackage{amsmath}
\usepackage{amssymb}
\usepackage{mathtools}
\usepackage{amsthm}

\usepackage[capitalize,noabbrev]{cleveref}

\theoremstyle{plain}

\theoremstyle{definition}

\theoremstyle{remark}

\input{template_iclr25/math_commands}

\usepackage{times}
\usepackage{helvet}
\usepackage{courier}
\usepackage{xcolor}
\usepackage{marvosym}
\usepackage{ifsym}

%% file: template_iclr25/math_commands.tex
\usepackage{amsmath,amsfonts,bm}

\def\eqref#1{equation~\ref{#1}}

\def\1{\bm{1}}

\def\ve{{\bm{e}}}

\def\vh{{\bm{h}}}

\def\vv{{\bm{v}}}

\def\vx{{\bm{x}}}

\DeclareMathAlphabet{\mathsfit}{\encodingdefault}{\sfdefault}{m}{sl}
\SetMathAlphabet{\mathsfit}{bold}{\encodingdefault}{\sfdefault}{bx}{n}

\def\sB{{\mathbb{B}}}

\def\sV{{\mathbb{V}}}



%% file: sections/0-author.tex
\author{
    Ruiyang Ma\equalcontrib\textsuperscript{\rm 1},
    Yunhao Zhou\equalcontrib\textsuperscript{\rm 2,3},
    Yipeng Wang\textsuperscript{\rm 4}, 
    Yi Liu\textsuperscript{\rm 2}, 
    Zhengyuan Shi\textsuperscript{\rm 2}, 
    Ziyang Zheng\textsuperscript{\rm 2}, \\
    Kexin Chen\textsuperscript{\rm 4}, 
    Zhiqiang He\textsuperscript{\rm 4}, 
    Lingwei Yan\textsuperscript{\rm 4}, 
    Gang Chen\textsuperscript{\rm 4}, 
    Qiang Xu\textsuperscript{\rm 2}, 
    Guojie Luo\textsuperscript{\rm 1, \Letter}
}
\affiliations{
    \textsuperscript{\rm 1}Peking University, 
    \textsuperscript{\rm 2}The Chinese University of Hong Kong, 
    \textsuperscript{\rm 3}Shanghai Jiao Tong University, \\
    \textsuperscript{\rm 4}Nanjing University of Aeronautics and Astronautics
    
}

%% file: sections/0-abstract.tex
\begin{abstract}

There is a growing body of work on using Graph Neural Networks (GNNs) to learn representations of circuits, focusing primarily on their static characteristics. However, these models fail to capture circuit runtime behavior, which is crucial for tasks like circuit verification and optimization. To address this limitation, we introduce DR-GNN (DynamicRTL-GNN), a novel approach that learns RTL circuit representations by incorporating both static structures and multi-cycle execution behaviors. DR-GNN leverages an operator-level Control Data Flow Graph (CDFG) to represent Register Transfer Level (RTL) circuits, enabling the model to capture dynamic dependencies and runtime execution. To train and evaluate DR-GNN, we build the first comprehensive dynamic circuit dataset, comprising over 6,300 Verilog designs and 63,000 simulation traces. Our results demonstrate that DR-GNN outperforms existing models in branch hit prediction and toggle rate prediction. Furthermore, its learned representations transfer effectively to related dynamic circuit tasks, achieving strong performance in power estimation and assertion prediction. 


\end{abstract}

%% file: sections/1-introduction.tex
\section{Introduction}

As circuit complexity increases, traditional algorithms for circuit design and optimization are increasingly challenged to meet the pressing demands for time and cost efficiency.
New models, representations, and methodologies for circuit understanding are urgently needed to advance hardware design and electronic design automation (EDA) research, enabling engineers to create efficient hardware solutions.

Recent works have begun to solve many canonical tasks in hardware development with deep learning methods~\cite{lcm_survey, eda_gnn_survey}. By training on extensive circuit data, these models have demonstrated the potential to understand the static feature of circuits and outperform traditional methods on some prediction tasks, including circuit quality (performance, power, area) estimation~\cite{sengupta2022good, master_rtl, lopera2021rtl}, combinational functionality representation~\cite{fgnn, deepgate_1, deepgate_2} and so on.

Despite these notable achievements, it has been observed that these models struggle with tasks that require in-depth analysis of circuit designs, especially those involving the dynamic behavior of circuits such as hardware verification, dynamic power estimation and so on~\cite{deepseq, design2vec}. The main reason is that current models exclusively depend on the static information of circuits (\ie hardware source code, netlist) for training. As a result, these models are limited to learning only structural or semantic information about the circuits.

However, the \emph{dynamic behavior} of circuits is equally important for understanding circuits and facilitating dynamic-related downstream tasks. These behaviors can reveal complex dependencies and interactions that are unapparent in static representation, thereby significantly enhancing the quality of circuit representation.
In this study, we diverge from traditional approaches  that focus on learning static representations of circuits. In an innovative endeavor, we train the model to learn circuit representations based on their dynamic behaviors. The dynamic behavior refers to the circuit's multi-cycle execution status under specific inputs. We aim to predict various circuit behavior features under specific input sequences, such as branch and assertion hits, variable toggle rate, dynamic power and so on.



We design DR-GNN (\textbf{D}ynamic\textbf{R}TL-GNN) to learn the dynamic behavior of circuits. DR-GNN is distinguished by two aspects. First, we construct the Graph Neural Network (GNN) based on the operator-level Control Data Flow Graph (CDFG) of Register Transfer Level (RTL) circuit design. The CDFG provides a high-level and concise representation of the dynamic functionality of the circuit, making it an ideal foundation for learning dynamic behaviors.
Second, we develop a circuit dynamic behavior-aware GNN propagation mechanism that integrates specific circuit features into the model. 
We design semi-decoupled aggregators tailored for numerous RTL operators and introduce a position-aware operator messaging mechanism in heterogeneous graph transformer~\cite{hgt}. This enables DR-GNN to capture both the structural and dynamic characteristics of circuits.


\begin{figure*}[ht]
    \centering
    \setlength\abovecaptionskip{2pt}
    \setlength\belowcaptionskip{-8pt}
    \includegraphics[width=0.9\linewidth]{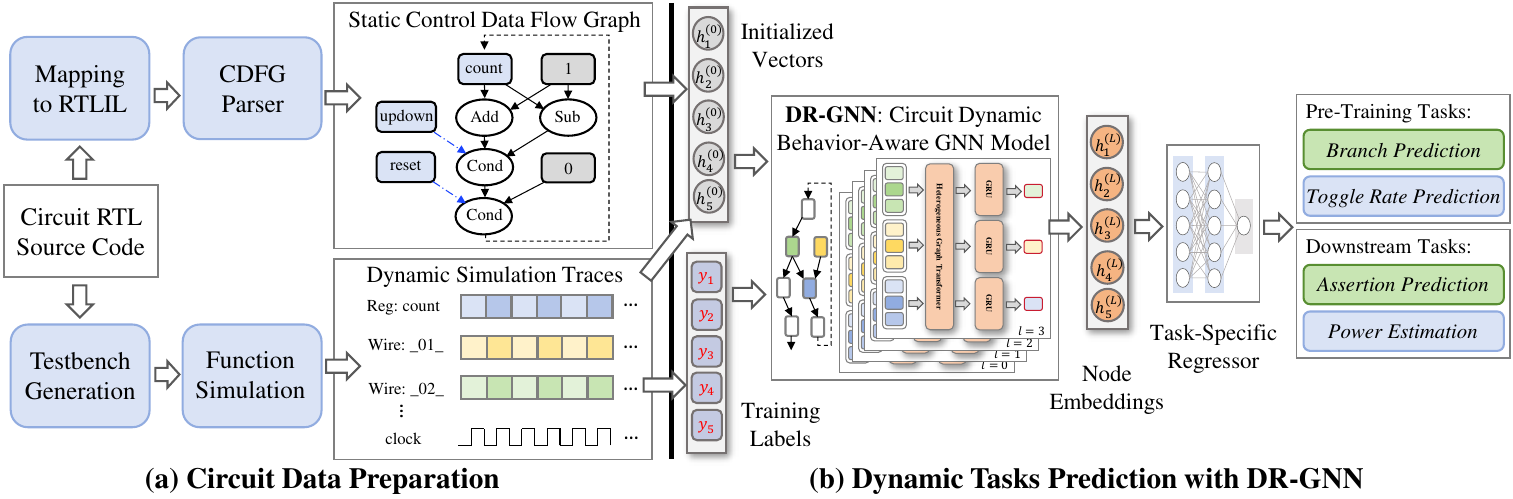}
    \caption{The overview of DynamicRTL. The framework prepares circuit training data, including CDFGs and simulation traces, and employs pre-training tasks to train the model. The learned representations are then used for downstream tasks.}
    \label{fig:overview}
\end{figure*}

We collect a comprehensive Verilog dataset and employ branch execution and variable toggle information as training supervision.
DR-GNN achieves outstanding performance in these pre-training tasks, significantly surpassing models based solely on circuit structural and semantic learning. 
Moreover, DR-GNN demonstrates effective transfer learning on downstream tasks including power estimation and assertion prediction, which highlights its ability to generate useful circuit representations for tasks related to circuit dynamic behaviors.
To our knowledge, DR-GNN is the first unified model that captures the dynamic behavior of hardware designs, setting the stage for teaching neural network models to better understand how circuits execute.

In summary, this paper makes the following contributions:
\begin{itemize}
    \item We introduce the problem of learning circuit dynamic representation. We create the first dynamic circuit dataset, comprising over 6,300 diverse Verilog designs, along with their CDFGs and 63,000 simulation traces. 
    \item We design the DR-GNN model
    which is built on the circuit operator-level CDFG and captures the dynamic functionality of the circuits.
    \item We obtain the first unified representation for circuit dynamic behaviors and achieve the state-of-the-art performance with an average accuracy of 94.80\% in branch hit prediction and 94.25\% in toggle rate prediction.
    \item We show that the pre-trained DR-GNN representations are useful for transfer learning, achieving an average accuracy of 79.60\% in power estimation and 87.03\% in assertion prediction.
\end{itemize}

The dataset, source code and supplementary materials are available at \url{github.com/magicyang1573/DynamicRTL}.


%% file: sections/2-background.tex
\section{Preliminary and DynamicRTL}


Figure~\ref{fig:overview} presents an overview of DynamicRTL. In this section, we discuss key concepts in our workflow. We introduce our method for representing the RTL circuit source code as a control data flow graph. Then, we explain the concept of dynamic behavior in circuits. Finally, we outline the tools used to extract these information from circuit designs.

\textbf{RTL Code.}
\textit{Register transfer level} (RTL) is an abstraction level in digital circuit designs. A digital circuit is composed of combinational logic (computing operators, branch controls) and sequential logic (registers). 
The assignment in RTL represents a physical connection between wires or registers. A wire is a conductive path for signal transmission, while a register is a small storage element that holds data temporarily during the execution of a digital circuit.
RTL is a representation that focuses on the flow of data between registers and the operations performed on those data. The register transferring data flow is commonly written in \cmd{always} blocks, which means that the logic is executed during every clock cycle. This is analogous to a software program written in a \cmd{while} loop, where the same code is repeatedly executed and the state continuously changes. 

\begin{figure}[t]
    \centering
    \setlength\abovecaptionskip{3pt}
    \setlength\belowcaptionskip{-5pt}
    \includegraphics[width=1.0\linewidth]{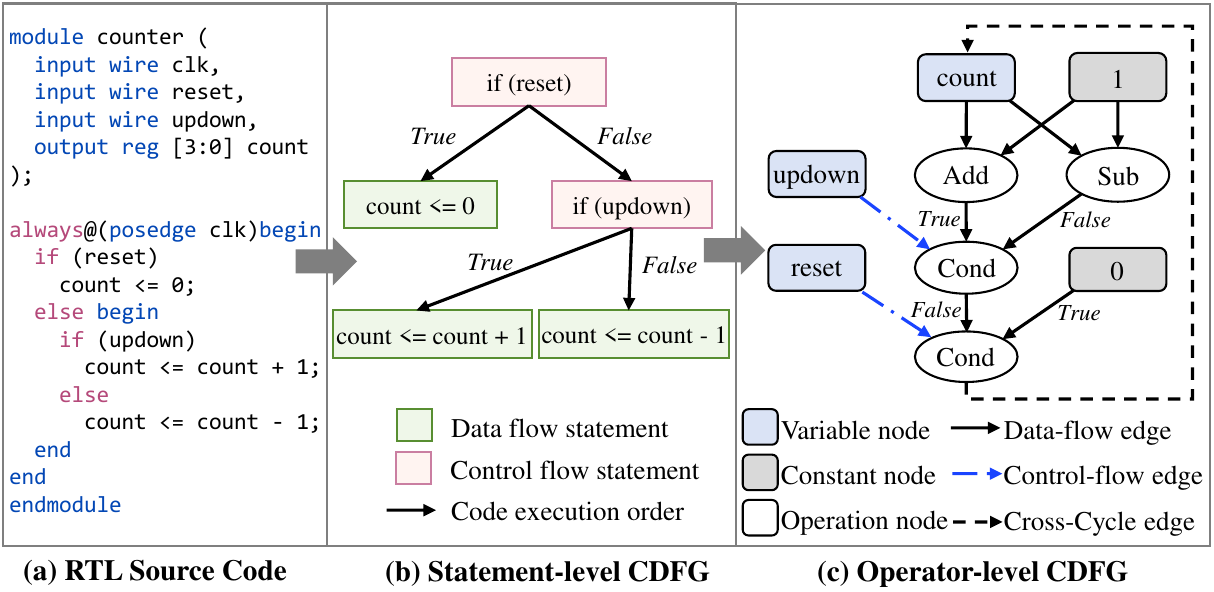}
    \caption{Comparison between statement-level CDFG and operator-level CDFG. In Verilog grammar, the operator \cmd{<=} denotes a non-blocking assignment to register.}
    \label{fig:cdfg}
\end{figure}

\textbf{Control Data Flow Graph.}
A common way to understand the function of RTL is \textit{control data flow graph} (CDFG). Figure~\ref{fig:cdfg} shows an example of RTL code and its CDFG. 
Previous work, Design2Vec~\cite{design2vec}, uses a statement-level CDFG to learn semantic representation of circuits, where each node represents an RTL statement. However, this CDFG only reflects the execution of circuit code from a software perspective, which fails to reflect the actual data flow within circuits. 

To better represent the dynamic behavior of circuits, we propose the operator-level CDFG as the circuit graph structure.
In the operator-level CDFG, directed edges indicate data or control flows between nodes. In sequential circuits, registers store values to the next clock cycle, therefore the in-edges of registers signify data flow across clock cycles. 


The CDFG comprises three node types: variable nodes, constant nodes and operator nodes.
\textit{Variable nodes} represent variables that have dynamic values, such as wires and registers. Input wires are special variable nodes whose values depend on the external drive. 
\textit{Constant nodes} represent unchanging values throughout clock cycles. 
\textit{Operator nodes} represent operations on variables in data flow. 
The \cmd{condition} node is a unique operator node, which acts as a multiplexer and uses select signals to control data flow from different channels.
The nodes in CDFG can be multi-bit, enabling a higher abstraction level and a more effective way to represent RTL behaviors compared with single-bit circuit graphs~\cite{master_rtl, deepgate_1}. For more details on circuit CDFG, refer to Appendix~\ref{ap:cdfg} in supplementary material.

\begin{figure*}[t]
\centering 
\includegraphics[width=0.8\textwidth]{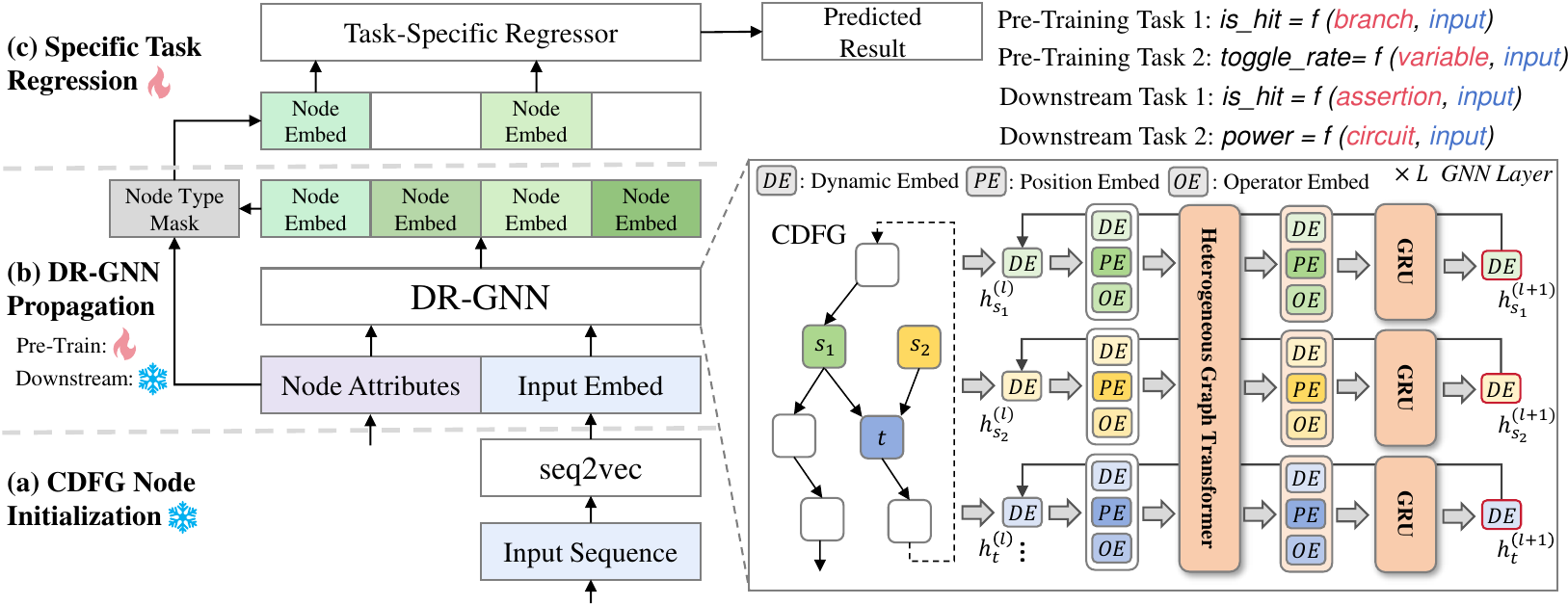} 
\setlength\abovecaptionskip{0pt}
\setlength\belowcaptionskip{-5pt}
\caption{The overview of circuit dynamic behavior-aware DR-GNN model.}
\label{fig:model}
\end{figure*}

\textbf{Circuit Dynamic Behavior.}
The dynamic behavior of a circuit refers to the changes in wire and register values during execution. Input signals drive these changes based on the combinational logic. Some values are stored in registers to be used in the next clock cycle. Given different input sequences, the circuit may exhibit significantly different behaviors, which has not been adequately addressed in prior studies on circuit representation.
In our work, we consider both sequential and combinational features of the circuit. We take the values of all wires and registers into account when assessing dynamic behavior. These values can be easily obtained through hardware simulation. 
We acquire the supervision from simulation traces and use branch prediction and variable toggle rate prediction as pre-training tasks, which will be detailed in Section~\ref{sec:pretrain_task} . 
After pre-training the model, we transfer the learned representation to downstream circuit dynamic tasks such as power estimation and assertion prediction, which will be introduced in Section~\ref{sec:exp-power} and \ref{sec:exp-assert}.

\textbf{Tools.}
To construct the CDFG, we first convert the source RTL code into RTL intermediate language (RTLIL) using EDA tool Yosys~\cite{yosys}. This conversion results in a functionally equivalent description, consisting only of assign statements and register transferring. Next, we use the Stagira Verilog parser~\cite{stagira} to parse the RTLIL and generate an \textit{abstract syntax tree} (AST). Finally, we traverse the AST to create circuit CDFG. To collect the dynamic traces of circuits, we use a commercial hardware simulator and capture the value of wires and registers in each clock cycle.

%% file: sections/4-model.tex

\section{DR-GNN: Circuit Dynamic Behavior-Aware GNN}

\subsection{Overview}
DR-GNN takes as input the CDFG and input sequence $I$. The circuit has branch set $\sB$ and variable set $\sV$. We use branch hit prediction and variable toggle rate prediction as the pre-training tasks. The objective of DR-GNN is to predict the branch hit probability $is\_hit(I, \sB)$ that input $I$ hit branches, and the toggle rate $toggle\_rate(I, \sV)$ that input $I$ trigger variable transitions. 

Figure~\ref{fig:model} presents the architecture of DR-GNN. The model is constructed based on the circuit CDFG. The embedding of each node represents its dynamic behavior.
DR-GNN operates in three stages:
Firstly, the initial dynamic embedding for each node is encoded based on its attribute and functionality (Section~\ref{sec:node_init}). 
Secondly, DR-GNN updates the node embeddings with circuit dynamic behavior-aware propagation (Section~\ref{sec:aggregation}).
Thirdly, the final embeddings of each node are used for various prediction tasks related to circuit dynamic behaviors (Section ~\ref{sec:pretrain_task}).

\subsection{CDFG Node Initialization}
\label{sec:node_init}

In DR-GNN model, node embeddings represent the dynamic behaviors. Before the GNN propagating, we encode some known dynamic behaviors as the initial embedding of CDFG nodes. 
The nodes are initialized based on three categories.  

\textbf{Input Sequence Encoder.} We encode the input sequence as the initial embedding of input nodes. Considering many circuit operators function at binary level, we design a binary input encoder, which involves two steps: embedding the value in each clock cycle, and integrating these embeddings into a sequence embedding.
We treat the value in each clock cycle as a binary vector and embed it with linear projection, similar to prior work on numerical representation~\citep{number_repre}. We define a learnable vector for each bit position, and sum these vectors element-wise, modulated by the value of corresponding bit. 
For the learnable vector $\ve_i$, for each bit and the $n$ bit value $v = \{b_1, b_2,..., b_n\}$, we compute $\vv_{emb} = \sum_{i=1}^n b_i\ve_i$. 
Then we use gated recurrent unit (GRU) to embed values into a sequence embedding and use that as the initial embedding of input nodes. 
We train the input sequence encoder independently before training the whole model. We use a sequence reconstruction task to simultaneously train an encoder and a decoder. 
After training, we freeze the encoder weight and use it as our input sequence encoder in the model.


\textbf{Constant Value Encoder.} The constant node can be seen as an input node that keeps the same value in each clock cycle. In order to ensure the embedding consistency of each node and reduce the complexity of GNN learning, we employ the same method as input encoder to encode the constant value. 

\textbf{Initialization of Operator Nodes.} The embedding of operator nodes represents the dynamic behavior of their calculated results in each clock cycle. Since we cannot predict these data before simulation, we initialize these embeddings to all zeros, indicating the unknown. 



\subsection{Circuit Dynamic Behavior-Aware Propagation}
\label{sec:aggregation}
Circuit signals are processed through hardware components, which correspond to the operator nodes in CDFG. We use the propagation in GNN to represent the dynamic behaviors.
Since the CDFG is a directed heterogeneous graph, we design the propagation mechanism referring to the heterogeneous graph transformer (HGT)~\citep{hgt}. 
However, there are challenges arising from the unique characteristics of circuit CDFGs that require tailored approaches for current HGT.

\textbf{Semi-Decoupled Operator Aggregation.} 
Circuit CDFGs contain diverse operators. 
Previous works on circuit representation focus mainly on netlist with limited operator types, which usually allocate a separate aggregator for each gate type to represent circuit functionalities~\cite{deepgate_1, fgnn}. 
These approaches are insufficient for representing RTL-level CDFG, which includes over 30 operators. In this case, using separate aggregators for each operator is impractical, as it would make the model overly complex and difficult to train. A potential solution is to use a shared aggregator combined with an attention-based messaging mechanism to differentiate operator functions. However, it is challenging to rely solely on a shared aggregator to accurately represent such a diverse array of operators. 

Therefore, we propose semi-decoupled aggregation mechanism to represent operators in circuit CDFG. 
Operators are categorized into five groups based on their characteristics and importance, as shown in Table~\ref{tab:aggregator}. 
Operators within each group share aggregation weights and use attention-based message mechanism to represent different functions. 
The grouping is based on operand number. The unary and binary groups include basic combinational operators. The multary group includes part-select and concat operators.
Additionally, two key operators are given individual aggregators: the condition operator, which handles branch selection in combinational logic, and the register variable, which stores data across clock cycles in sequential logic.

\begin{table}
    \setlength\abovecaptionskip{5pt}
    \centering
    \footnotesize
    \begin{tabular}{c|c|c}
        \toprule
        \noalign{\vspace{-2pt}}
       \textbf{Group} & \textbf{Num} & \textbf{Description} \\\hline
         Unary & 7 & Operators with one operand \\\hline
         Binary & 20 & Operators with two operands \\\hline
         Multary & 2 & Operators with more than two operands \\\hline
         Cond. & 1 & Condition operator for branch selection \\\hline
         Reg. & 1 & Register variable treated as operator \\
        \noalign{\vspace{-2pt}}
        \bottomrule
    \end{tabular}
    \caption{Semi-decoupled aggregator groups for operators.}
    \vspace{-10pt}
    \label{tab:aggregator}
\end{table}


\textbf{Position-Aware Operator Messaging.}
To represent different functions of operators within each aggregation group, we use attention mechanism in message passing. It is crucial to consider necessary information for the attention to effectively capture circuit dynamic behaviors. 
The first necessary information is the operator type, which directly affects the computing results. 
Additionally, some operators are non-commutative. For instance, the subtraction operator yields different results for \cmd{a-b} and \cmd{b-a}. Therefore, it is also essential to consider the relative position of operand nodes to the operator when aggregating embeddings. Traditional position encodings in graph transformer 
are insufficient to represent such operand-to-operator positional relationships~\cite{gt_recipe}. To address this, we incorporate operand position encoding and design a position-aware operator messaging mechanism in HGT to represent circuit dynamic behaviors.


Consider an operator node $t$ with dynamic embedding $\vh_t^{(l)}$ in the $l$-th GNN layer. Node $t$ has multiple source nodes $s_1, s_2, ..., s_m$, each with dynamic embeddings
$\vh_{s_1}^{(l)}, \vh_{s_2}^{(l)}, ... , \vh_{s_m}^{(l)}$. 
The type of the operator node is denoted as $op[t]$. The position of each source node $s$ relative to $t$ is denoted as $pos[s, t]$, indicating the order of operand $s$ in the expression of operator $t$. 
We use rotary position embedding to encode these positions.
The input to the heterogeneous graph transformer is then constructed by concatenating the source node position embedding, the operator type embedding, and the dynamic embedding of the source node.
\begin{eqnarray}
\vx_{s,t}^{(l)} = \mathrm{Concat}(\vh_s^{(l)}, \mathrm{Embed}(pos[s, t]), \mathrm{Embed}(op[t]))
\end{eqnarray}
In the heterogeneous graph transformer, we map the embedding of source node $s$ and target node $t$ into Query and Key vectors. The concatenation of the Query and Key vectors serves as the input to the attention mechanism. The attention weight from node $s$ to $t$ is calculated by:
\begin{eqnarray}
w_{s,t}^{(l)} = \mathrm{MLP}(\mathrm{Concat}(\mathrm{Q\text{-}linear}(\vx_{s,t}^{(l)}), \mathrm{K\text{-}linear}(\vh_t^{(l)})))
\end{eqnarray}
For each source node $s$ connected to node $t$, the normalized attention weights are computed using the softmax function:
\begin{eqnarray}
\{a_{s_1,t}^{(l)}, ... ,a_{s_m,t}^{(l)}\} = \mathrm{Softmax}(\{w_{s_1,t}^{(l)}, ... ,w_{s_m,t}^{(l)}\})
\end{eqnarray}
The message from $s$ to $t$ is represented by a Value vector:
\begin{eqnarray}
message_{s, t}^{(l)} = \mathrm{V\text{-}linear}(\vx_{s,t}^{(l)})
\end{eqnarray}
Finally, the messages from all source nodes are aggregated using the attention weights to obtain the aggregated information for node $t$:
\begin{eqnarray}
aggr_t^{(l)} = \sum\nolimits_{i=1}^m message_{s_i, t}^{(l)} \cdot a_{s_i,t}^{(l)}
\end{eqnarray}

\textbf{Node Embedding Update.}
We use GRU to update the embedding of target node $t$, where $aggr_t^{(l)}$ is the aggregation information as the GRU input and $\vh_t^{(l)}$ is the past state of GRU. The output $\vh_t^{(l+1)}$ serve as the dynamic embedding for the subsequent $(l{+}1)$-th GNN layer. 
\begin{eqnarray}
\vh_t^{(l+1)} = \mathrm{GRU}(aggr_t^{(l)}, \vh_t^{(l)})
\end{eqnarray}
By stacking $L$ layers, we obtain the node representations $\vh^{(L)}$ for the entire graph. These representations can be used for prediction of the pre-training and downstream tasks.

\subsection{Pre-training Tasks}
\label{sec:pretrain_task}
Task 1 predicts the branch hit probability. 
Hardware branches, represented by conditional statements like \cmd{if} or \cmd{case} in RTL code, capture abundant dynamic circuit behaviors and are key coverage metrics in hardware verification.
We identify nodes serving as select signals for the \cmd{condition} node to form the branch set $\sB$. For each branch node $b$, we read out its embedding $h_b^{(L)}$ and pass it through a multi-layer perceptron to predict the hit probability.
\begin{eqnarray}
\hat{P_{b}} = \mathrm{MLP_{branch}}(\vh_b^{(L)}), \quad b \in \sB
\end{eqnarray}
Task 2 predicts the variable toggle rate. In synchronous sequential circuits, a toggle occurs when the value of a variable changes, typically on the edge of the clock signal. 
The toggle rate of a variable refers to the number of toggles divided by the total number of clock cycles. This metric indicates how frequently a variable changes and is crucial for downstream tasks like power estimation.
We select nodes representing variables to form the variable set $\sV$, and use each node's embedding $h_v^{(L)}$ to predict the toggle rate. 
\begin{eqnarray}
\hat{R_{v}} = \mathrm{MLP_{toggle}}(\vh_v^{(L)}), \quad v \in \sV
\end{eqnarray}
We use multiple supervisions in pre-training to learn a comprehensive circuit dynamic representation, which captures diverse behaviors and improves generalization. In Section~\ref{sec:exp-power} and \ref{sec:exp-assert}, we will show that abundant supervisions enhance performance in each downstream task.



%% file: sections/5-experiment.tex
\section{Experiments}
\input{tables/main-exp}

\subsection{Dataset Preparation}
\textbf{Circuit Designs Collection. } 
To train the DR-GNN model, we construct the first dynamic circuit dataset, which consists of around 6,300 different circuit designs and 63,000 circuit simulation traces.
We collect existing open-source Verilog datasets~\citep{mgverilog, VeriGen}. 
To further expand our collection, we search GitHub using keywords such as ``Verilog'', ``RTL'', and ``circuit'' and scrape Verilog files from relevant repositories.  
The resulting collection spans diverse practical circuit designs, including RISC-V components, SoC components, accelerator modules, custom IP blocks and so on.
During the preprocessing stage, we filter out designs with syntax errors, synthesis or simulation failures. We retain only sequential circuits in the dataset.


The collected designs range in size from 10 to over 500 CDFG nodes. The scale of these circuits is comparable to prior circuit representation works~\cite{fgnn, deepgate_2, master_rtl}. A detailed introduction to the dataset is provided in Appendix~\ref{ap:dataset} in supplementary material. 
Additionally, we also evaluate the generalizability of DR-GNN on larger-scale circuits ranging from 1k to 10k CDFG nodes in downstream tasks, which are detailed in Section~\ref{sec:exp-power}.



\textbf{Simulation Traces Collection. }
We use commercial simulator to collect the traces. Random input patterns are generated, and each circuit is simulated with 10 traces. During simulation, all internal variable values are recorded at each clock cycle. When generating testbench, we automatically identify special signals such as the reset signal. The reset signal is active only at the beginning of simulation and remains inactive thereafter, which ensures the simulation traces capture a comprehensive range of circuit behaviors.

\subsection{Experimental Settings}


We split our dataset by designs, using 80\% of the designs for training, 10\% for validation and 10\% for testing.
This ensures the designs used for testing are never seen by the model during training and validation, providing a reliable measure of its generalization ability. 
We use binary cross-entropy loss for branch hit prediction and mean squared error loss for toggle rate prediction, combining them to form the final loss. 
Experiment is conducted on an A800 GPU.
A detailed introduction to the training process and model hyperparameters is provided in Appendix~\ref{ap:model_detail} in supplementary material.

\subsection{Pre-training Tasks}
\label{sec:exp-pretrain}

\textbf{Baseline Comparison.}
We evaluate DR-GNN model on two pre-training tasks using input sequences of varying lengths. 
Since these tasks focus on branches and variables at RTL level, which are challenging to map to synthesized netlist circuits, we compare DR-GNN with two RTL-level circuit representation frameworks. 
HGVC~\cite{sengupta2022good} extracts CDFG features (e.g., node type, bit width) as initial node embeddings and applies GNNs for prediction tasks. We apply GCN~\cite{gcn}, GAT~\cite{gat}, Gated GCN~\cite{gated_gcn} and GATv2~\cite{gatv2} as baselines in the experiments. For a fair comparison, we incorporate dynamic information into HGVC and use the same initialization for input node embeddings as in DR-GNN. 
Design2Vec~\cite{design2vec} constructs the GNN based on statement-level CDFG and learns semantic embeddings. Because its nodes represent statements rather than variables, we evaluate it only on branch hit prediction, consistent with its original study.

\begin{figure}[h]
    \centering
    \setlength\abovecaptionskip{5pt}
    \begin{subfigure}{0.24\textwidth}
        \centering
        \includegraphics[width=\textwidth]{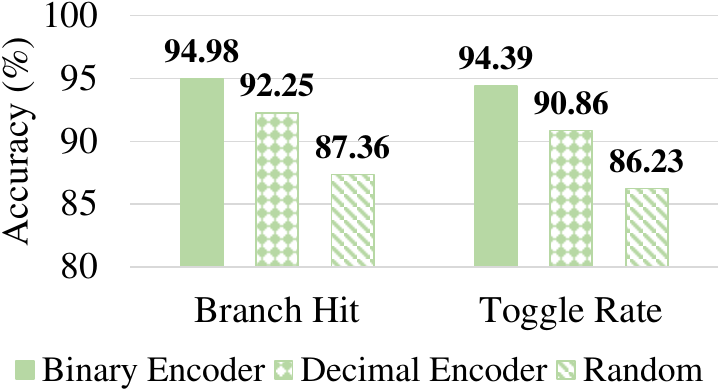}
        \caption{Influence of input encoders.}
        \label{fig:ablation_input}
    \end{subfigure}%
    \begin{subfigure}{0.24\textwidth}
        \centering
        \includegraphics[width=\textwidth]{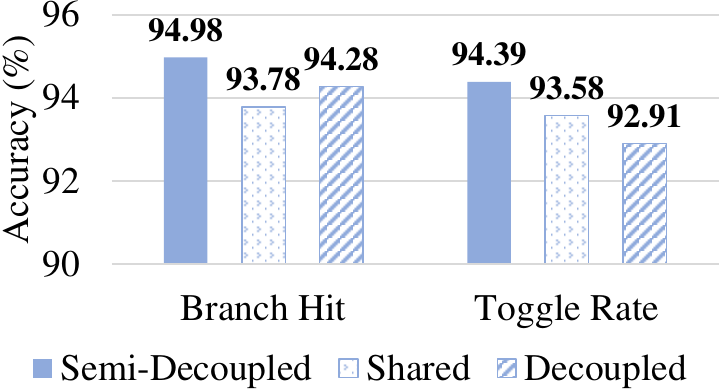}
        \caption{Influence of aggregators.}
        \label{fig:ablation_aggr}
    \end{subfigure}
    \caption{Ablation Study. Branch hit and toggle rate prediction on different input encoding and aggregation mechanisms.}
    \label{fig:ablation_1}
\end{figure}

Table~\ref{tab:model_comparison} presents the experiment results. All models are trained for 5 times and the mean and standard variance of test performance are reported. 
The DR-GNN model achieves accuracy of about 94\% in both branch hit prediction and toggle rate prediction
\footnote{We define toggle rate accuracy as $1-|\hat{R_{v}}-R_{v}|$, consistent with the approach in prior study~\cite{deepseq}.} 
across different input sequence lengths, outperforming all baselines.
\textbf{Ablation Study of Input Encoding.}
The dynamic behavior of circuits is determined by the input sequence, which we encode in DR-GNN using a binary input encoder. We compare DR-GNN with two ablation variants, as shown in Figure~\ref{fig:ablation_input}.
In the random encoding, we ignore the input sequence and provide random value as the model's input, excluding dynamic information from the model.
In the decimal encoding, we treat the input sequence as decimal values rather than a binary vectors.
Experiments demonstrate that encoding the input sequence as a binary vector sequence allows the DR-GNN model to achieve the best performance. 

\textbf{Ablation Study of Aggregator.}
To handle the diversity of operator types in CDFG while avoiding overly large models, we design a semi-decoupled aggregation mechanism. We evaluate two ablation variants, as shown in Figure~\ref{fig:ablation_aggr}.
In the decoupled aggregation mechanism, each operation type is assigned an independent aggregator. In the shared aggregation mechanism, a single aggregator is used for all operators, with their different functions represented through an attention mechanism. 
The results demonstrate that the semi-decoupled aggregation achieves the best performance.

\begin{figure}[h]
    \centering
    \includegraphics[width=0.4\textwidth]{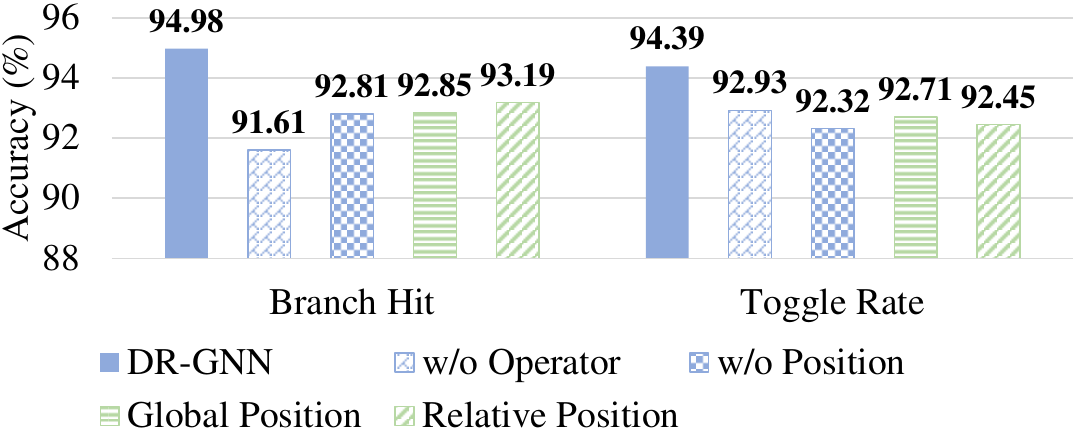}
    \caption{Ablation Study. Branch hit and toggle rate prediction accuracy on different messaging mechanisms.}
    \label{fig:ablation_2}
\end{figure}

\textbf{Ablation Study of Messaging.}
We propose a position-aware operator messaging mechanism in DR-GNN, which incorporates both operator types and operand positions in the attention mechanism. 
Ablation studies compare the performance with variants that eliminate the operator type (w/o Operator) and the operand positions (w/o Position) from the attention mechanism.
Additionally, we compare our operand position embedding with two existing position encoding methods in graph transformer: global position encoding using Laplacian eigenvectors~\cite{global_pos} and relative position encoding using distance embedding~\cite{relative_pos}. These position encodings are incorporated into the initial node embeddings.
The results in Figure~\ref{fig:ablation_2} show that the position-aware messaging achieves the best performance.

\begin{figure}[h]
    \centering
    \setlength\abovecaptionskip{5pt}
    \begin{subfigure}{0.24\textwidth}
        \centering
        \includegraphics[width=\textwidth]{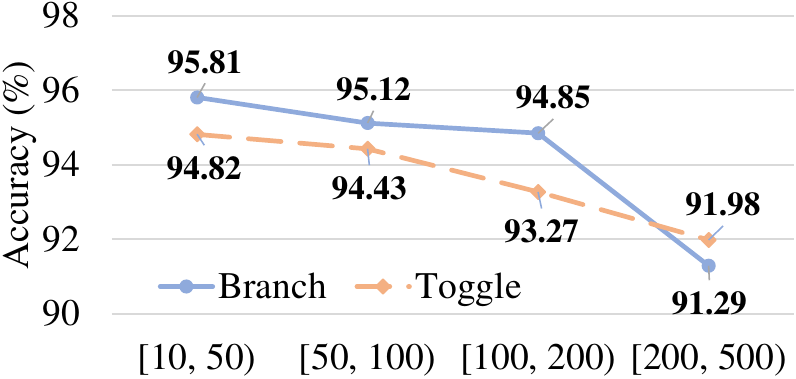}
        \caption{Influence of CDFG nodes.}
        \label{fig:subfig1}
    \end{subfigure}%
    \begin{subfigure}{0.24\textwidth}
        \centering
        \includegraphics[width=\textwidth]{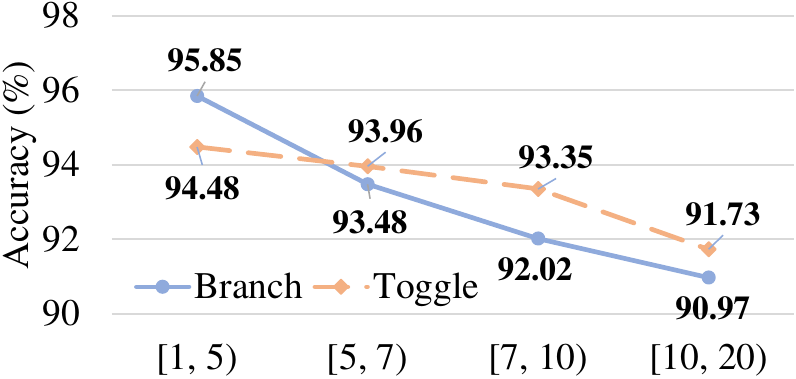}
        \caption{Influence of CDFG depth.}
        \label{fig:subfig2}
    \end{subfigure}
    \caption{Branch hit and toggle rate prediction on different circuit CDFG size and depth. }
    \label{fig:scale_exp}
\end{figure}

\textbf{Circuit Scale Variation Analysis.}
We evaluate the performance of DR-GNN on circuit designs of different sizes and layer depths, as presented in Figure~\ref{fig:scale_exp}. 
The circuit size is defined by the node count of the CDFG, and the layer depth is defined by the longest path from input nodes to output nodes in CDFG.
For designs with fewer than 50 nodes and 5 layers, our model achieves over 95\% branch hit prediction accuracy and more than 94\% toggle rate prediction accuracy. 
As design size increases, the complexity of understanding rises, while our model maintains good performance. 

%% file: tables/main-exp.tex
\begin{table*}[t]    
    \setlength\abovecaptionskip{5pt}
    \centering
    \fontsize{8}{9}\selectfont
    \begin{tabular}{l|*{4}{>{\centering\arraybackslash}p{1.45cm}}|*{4}{>{\centering\arraybackslash}p{1.45cm}}}
        \toprule
        \noalign{\vspace{-2pt}}
        \multicolumn{1}{c|}{Pretraining task} & \multicolumn{4}{c|}{Branch Hit Prediction (\%) } & \multicolumn{4}{c}{Toggle Rate Prediction (\%)} \\ 
        \cline{1-9}
        \noalign{\vspace{0.5pt}}
        \multicolumn{1}{c|}{Seq. length} & 10 & 20 & 30 & 50 & 10 & 20 & 30 & 50 \\ \hline
        \noalign{\vspace{0.5pt}}
        GCN & 81.75$\pm$0.58 & 82.24$\pm$0.52 & 82.68$\pm$0.78 & 82.97$\pm$0.59 & 81.88$\pm$0.51 & 81.15$\pm$0.35 & 80.99$\pm$0.58 & 80.93$\pm$0.47 \\ 
        GAT & 82.68$\pm$0.42 & 83.57$\pm$0.72 & 82.70$\pm$0.63 & 83.81$\pm$0.41 & 82.38$\pm$0.53 & 81.96$\pm$0.68 & 81.49$\pm$0.33 & 81.79$\pm$0.63 \\ 
        Gated GCN & 83.52$\pm$0.64 & 83.96$\pm$0.40 & 84.15$\pm$0.57 & 84.34$\pm$0.68 & 83.07$\pm$0.30 & 82.84$\pm$0.54 & 82.46$\pm$0.35 & 82.62$\pm$0.37 \\ 
        GATv2 & 85.23$\pm$0.71 & 85.60$\pm$0.76 & 85.72$\pm$0.56 & 86.16$\pm$0.78 & 85.25$\pm$0.37 & 85.11$\pm$0.46 & 84.37$\pm$0.38 & 84.55$\pm$0.51 \\ 
        Design2Vec & 82.03$\pm$0.88 & 86.78$\pm$0.65 & 72.81$\pm$1.13 & 69.33$\pm$0.91 & - & - & - & - \\ 
        \textbf{DR-GNN} & \textbf{94.69}$\pm$0.75 & \textbf{94.75}$\pm$0.64 & \textbf{94.98}$\pm$0.59 & \textbf{94.78}$\pm$0.69 & \textbf{93.74}$\pm$0.41 & \textbf{93.98}$\pm$0.47 & \textbf{94.39}$\pm$0.34 & \textbf{94.90}$\pm$0.36\\ 
        \noalign{\vspace{-2pt}}
        \bottomrule
    \end{tabular}
    \caption{Comparison of accuracy of different models for branch hit prediction and variable toggle rate prediction tasks.}
    \label{tab:model_comparison}
    \vspace{-12pt}
\end{table*}

%% file: sections/5.1-downstream-exp.tex
\section{Downstream Tasks}
This section evaluates the generalizability of the pre-trained DR-GNN model on two downstream tasks: power estimation and assertion prediction. For both tasks, we use the learned dynamic circuit representation as the circuit node feature and use that in the model for downstream tasks. These downstream tasks figure out whether DR-GNN effectively capture the dynamic behaviors of circuit and enhance the performance of dynamic-related circuit tasks.

\subsection{Evaluation on Power Estimation}
\label{sec:exp-power}

In circuit development, power estimation is time-consuming and challenging due to its strong dependence on circuit dynamic behaviors. 
We evaluate DR-GNN on RTL-level dynamic power estimation task. 
In our study, we use the learned dynamic representation, concatenated with node attributes including node type and width, as node embeddings for estimation. With DR-GNN weights frozen, we introduce an additional GAT based on CDFG to estimate power.

\textbf{Experimental Settings. }
For each design, we synthesize the RTL code using commercial synthesis tool with the NanGate 45nm technology library~\cite{nangate_45}. The dynamic power of the gate-level netlist under specific input is recorded as the ground-truth label.
We compare DR-GNN with two baselines. 
MasterRTL~\cite{master_rtl} models circuit as bit-level simple operator graph (SOG), which is close to netlist, and extracts toggle rate as node features. 
It then uses a tree-based model to estimate power. 
HGVC~\cite{sengupta2022good} models the circuit as CDFG, similar to ours, using node attributes (\eg type, width) as initial embeddings and a GAT for power prediction.
We train these models on our dataset, which is significantly larger than the dataset used in their original studies.
We use correlation coefficient (R), mean absolute percentage error (MAPE), and root relative square error (RRSE) to evaluate the accuracy between predicted value $\hat{y}$ and ground-truth $y$, as consistent with prior work~\cite{master_rtl}.
\begin{eqnarray}
\text{MAPE}= \dfrac{1}{n}\sum_{i=1}^{n}\dfrac{|y_i-\hat{y_i}|}{y_i}, \ \ \text{RRSE} = \sqrt{\dfrac{\sum_{i=1}^{n}{(y_i-\hat{y_i}})^2}{\sum_{i=1}^{n}{(y_i-\bar{y}})^2}}
\end{eqnarray}

\textbf{Power Estimation Results. }
Figure~\ref{fig:power_fig} presents the results. Using the pre-trained dynamic representation, our DR-GNN model achieves the best accuracy compared to MasterRTL and HGVC. This is because the dynamic embedding is trained on abundant circuit data and supervisions, which contain more information than manually extracted features. Moreover, our model outperforms MasterRTL without the need for synthesis into the SOG structure or running simulations to obtain toggle rates, which demonstrates the effectiveness of our dynamic representations. 
We also investigate the impact of using different pre-training supervisions on the power estimation performance. Experiment results show that using both types of supervision leads to better performance than using toggle supervision alone (DR-GNN$_{ts}$) or branch supervision alone (DR-GNN$_{bs}$).

\input{tables/power-exp}

\input{tables/exp-large-circuit-power}
\textbf{Power Estimation on Large Scale Circuits. }
Table~\ref{tab:large-power} presents the accuracy and runtime improvements achieved by DR-GNN on five large-scale circuit designs, compared to a netlist-level power estimation method DeepSeq2~\cite{deepseq2}, and the conventional power estimation flow using EDA tools.
For small-scale circuits, the netlist-level model exhibits higher accuracy. This is because the circuit netlist provides a more fine-grained representation of the circuit structure compared to the RTL-level CDFG.  However, this comes at the cost of a significantly larger graph structure in the netlist-level And-Inverter Graph (AIG). DeepSeq2 cannot handle the three larger-scale circuits due to memory constraint (illustrated as MO in the table). 

To conclude, DR-GNN has stronger applicability to large-scale circuits compared with netlist-level models. Moreover, with the dynamic circuit representation learned from large training datasets, DR-GNN significantly improves the power estimation accuracy at the RTL level compared to MasterRTL and HGVC, which enables more accurate power estimation results for circuits at the early design stage.

\subsection{Evaluation on Assertion Prediction}
\label{sec:exp-assert}
The assertion prediction task reveals what dynamic representation DR-GNN has learned in pre-training.
With DR-GNN weights frozen, a multi-layer perceptron is trained for each assertion. The model predicts whether a variable satisfies the assertion under given input. 
We formulate eight assertions. For single-variable assertions, we select all the possible circuit variables as prediction target. For dual-variable assertions, we target possible variable pairs.

\input{tables/assertion-exp}

Table~\ref{tab:exp-assert} presents the assertions and their prediction accuracies. The results indicate that the learned embeddings effectively predict some simple assertions. DR-GNN accurately assesses approximate value ranges for assertions like \cmd{v<4} and \cmd{v<16}. This capability is also evident in assertions requiring variable comparisons, such as assertion \cmd{v1$\neq$v2} and \cmd{v1$<$v2}. Besides, our bit-wise signal embedding approach enables the representation to support bit-level operations like bit-and (\cmd{$\&$}), bit-or (\text{ }\textbar \text{ }).
However, we also find limitations that DR-GNN struggles to predict exact values of variables. For assertion like \cmd{v$\neq$2} and \cmd{v$\neq$4}, the prediction accuracy is lower, which leaves room for improvement. 

Table~\ref{tab:exp-assert} also presents prediction performance of DR-GNN$_{ts}$ (trained only with toggle supervision) and DR-GNN$_{bs}$ (trained only with branch supervision). The original DR-GNN exhibits best performance, which demonstrates the benefits of both supervisions. It also proves that supplementing more useful supervisions could help learn a more comprehensive representation of circuit dynamic behaviors.

%% file: tables/power-exp.tex
\begin{figure}[h]
    \setlength\abovecaptionskip{5pt}
    \centering
    \begin{subfigure}[t]{0.43\linewidth}
        \centering
        \includegraphics[width=\textwidth]{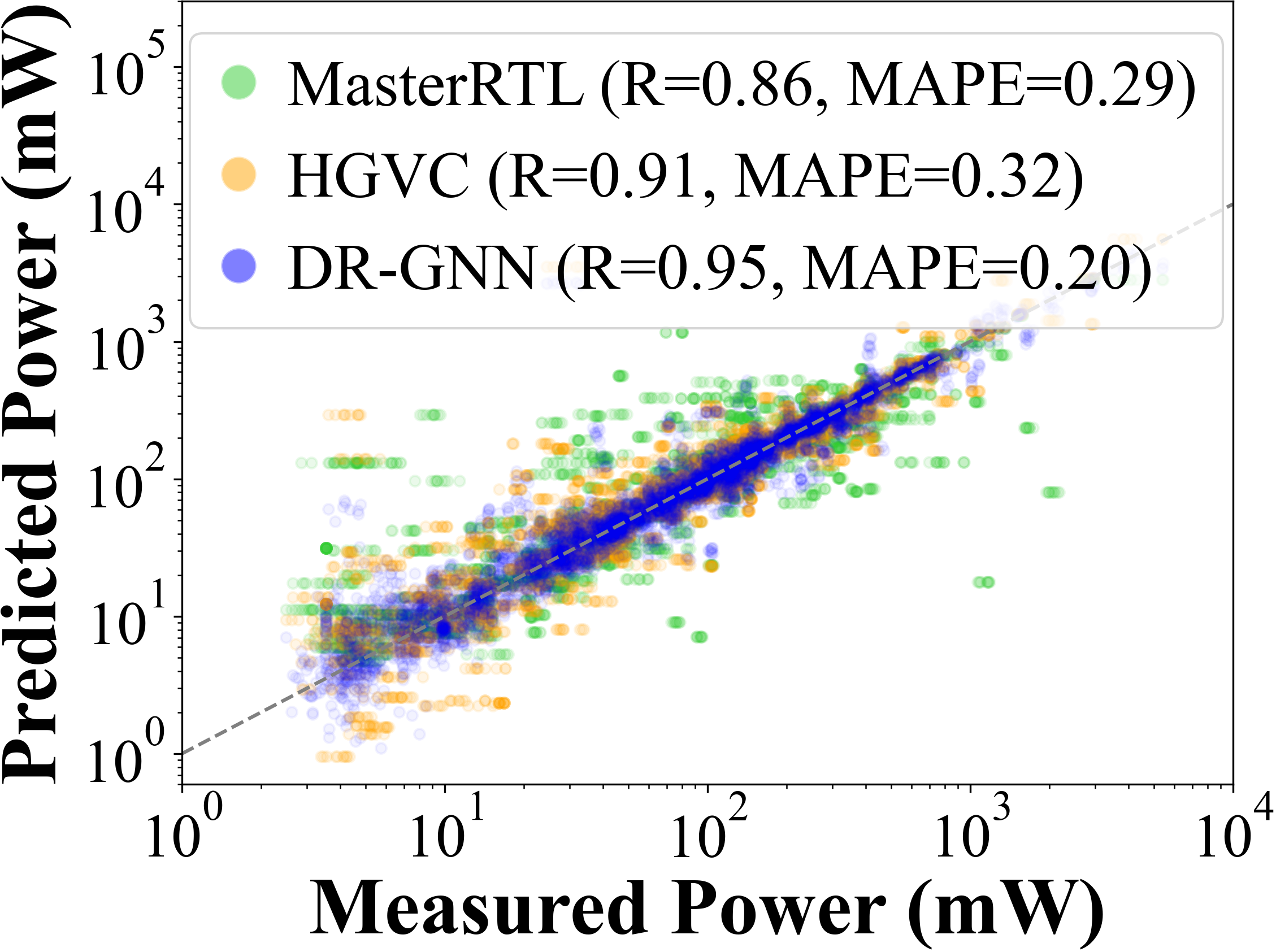}
        \caption{Predict vs ground-truth.}
    \end{subfigure}
    \hfill
    \begin{subfigure}[t]{0.52\linewidth}
        \centering
        \includegraphics[width=\textwidth]{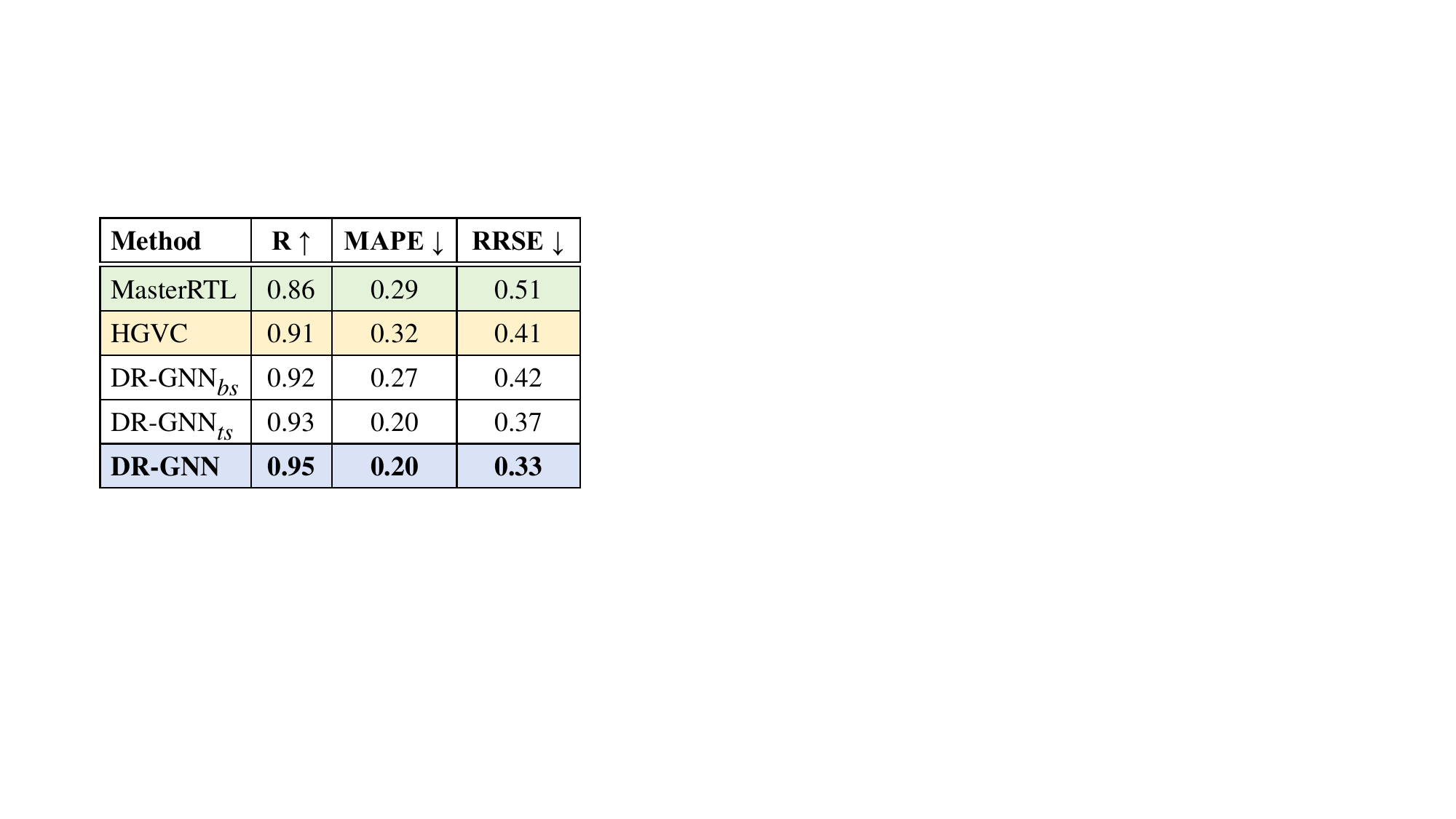}
        \caption{Accuracy comparison.}
    \end{subfigure}
    \caption{Power estimation comparison of MasterRTL, HGVC and DR-GNN with different supervisions. }
    \label{fig:power_fig}
\end{figure}

%% file: tables/exp-large-circuit-power.tex

\begin{table}[b]
    \setlength\abovecaptionskip{5pt}
    \renewcommand{\arraystretch}{1.1}
    \centering
    \setlength{\tabcolsep}{3pt}    
    \fontsize{8}{9}\selectfont
    \begin{tabular}{l|cc|cc|cc|c}
    \toprule
    \noalign{\vspace{-2pt}}
    \multirow{2}{*}{Design} & CDFG & AIG &
    \multicolumn{2}{c|}{DeepSeq2} & \multicolumn{2}{c|}{DR-GNN} & EDA \\
    \cline{4-7}
    & Nodes & Nodes & Error & Time (s) & Error & Time (s) & Time (s)\\
    \hline
    pcie\_ctrl & 0.9k & 22k & 0.08 & 14.8 & 0.15 & 0.17 & 540 \\
    dma\_sched & 1.9k & 43k & 0.09 & 27.2 & 0.14 & 0.29 & 1232 \\
    ysyx\_cpu & 6.0k & 120k & MO & MO & 0.19 & 0.61 & 2479 \\
    csc\_wl\_dec & 7.4k  & 185k & MO & MO & 0.18 & 0.73 & 2175\\
    ch\_ctrl & 8.3k & 304k & MO & MO & 0.23 & 0.96 & 2896 \\
    \noalign{\vspace{-2pt}}
    \bottomrule
    \end{tabular}
    \caption{Power estimation on 5 large-scale circuit designs.}
    \label{tab:large-power}
\end{table}

%% file: tables/assertion-exp.tex

\begin{table}[h]
    \setlength\abovecaptionskip{5pt}
    \centering
    \setlength{\tabcolsep}{3pt}    
    \fontsize{8}{9}\selectfont
    \begin{tabular}{l|ccc|c}
    \toprule
    \noalign{\vspace{-2pt}}
        Assertion & DR-GNN$_{ts}$ & DR-GNN$_{bs}$ & \textbf{DR-GNN} & Frequency \\
        \hline
        \noalign{\vspace{0.5pt}}
        \cmd{v < 4} & 82.37 & 81.76 & \textbf{86.98} & 56.78 \\ 
        \cmd{v < 16} & 81.74 & 82.57 & \textbf{87.35} & 61.52 \\ 
        \cmd{v $\neq$ 2} & 80.88 & 79.39 & \textbf{82.01} & 70.69 \\ 
        \cmd{v $\neq$ 4} & 80.59 & 76.38 & \textbf{81.36} & 73.98 \\ 
        \cmd{v1 $\neq$ v2} & 88.47 & 84.55 & \textbf{91.36} & 66.51 \\ 
        \cmd{v1 < v2} & 81.67 & 77.18 & \textbf{88.72} & 50.89 \\ 
        \cmd{v1\&v2$==$0} & 82.85 & 66.80 & \textbf{86.80} & 60.25 \\ 
        \cmd{v1|v2 $\neq$ 0} & 89.21 & 85.63 & \textbf{91.67} & 68.83 \\ 
        \noalign{\vspace{-2pt}}
        \bottomrule
    \end{tabular}
    \caption{Assertion prediction with pre-trained representation.}
    \label{tab:exp-assert}
\end{table}

%% file: sections/7-conclusion.tex
\section{Conclusion and Future Work}
This paper introduces DR-GNN model which learns dynamic representations for digital circuits. The model is pre-trained on the tasks of branch hit prediction and toggle rate prediction. It then demonstrates transfer learning capabilities on downstream tasks, such as power estimation and assertion prediction, including applications to large-scale circuit designs. For the first time, we showcase the ability of deep learning to capture complex temporal logic in sequential circuits, which facilitates a more abundant understanding of circuits and promotes a more efficient circuit design process.

%% file: appendix/appendix.tex
\newpage
\appendix
\onecolumn

\input{appendix/appendix-A}
\newpage

\input{appendix/appendix-B}
\newpage

\input{appendix/appendix-C}
\newpage

\input{sections/6-realted-work}

%% file: appendix/appendix-A.tex
\section{Model Details}
\label{ap:model_detail}

\subsection{Hyperparameters}
The hyperparameters for the DR-GNN model are detailed in Table~\ref{tab:hyperparameters}.

\renewcommand{\arraystretch}{1.5}

\begin{table}[h]
    \centering
    \begin{tabular}{|c|c|c|}
        \hline
        \multirow{7}{*}{DR-GNN} & input feature size & 1536 \\ \cline{2-3}
        & operation embedding size & 32 \\  \cline{2-3}
        & position embedding size & 32 \\  \cline{2-3}
        & hidden size & 1536 \\ \cline{2-3}
        & GNN layer & 20 \\ \cline{2-3}
        & GRU layer & 1 \\ \cline{2-3}
        & optimizer & adam \\\cline{2-3}
        & learning rate & 0.0001 \\ \hline
        \multirow{5}{*}{seq2vec} & input number size & 32 \\ \cline{2-3}
        & hidden size & 512 \\ \cline{2-3}
        & GRU layer & 3 \\ \cline{2-3}
        & optimizer & adam \\ \cline{2-3}
        & learning rate & 0.0001 \\ \hline
        \multirow{2}{*}{MLP for task prediction} & layer & 2 \\ \cline{2-3}
        & hidden size & 50 \\ \cline{2-3} \hline
        \multirow{2}{*}{GCN for power estimation} & layer & 5 \\ \cline{2-3}
        & hidden size & 1536 \\ \cline{2-3}
        \hline
    \end{tabular}
    \caption{Hyperparameters for models.}
    \label{tab:hyperparameters}
\end{table}

\subsection{Training and Inference Details}
The training and inference of DR-GNN are conducted on a single A800 GPU. The model is trained for 60 epochs using the Adam optimizer with a learning rate of 0.0001 and a batch size of 32. Our dataset comprises approximately 6,300 circuit designs, each with 10 simulation traces, resulting in around 63,000 data entries. We allocate 80\% of the circuits for training. 
DR-GNN requires approximately 40 minutes per epoch for training on pre-training tasks, with an inference time of about 0.7 seconds per batch. When the DR-GNN weights are frozen and the GAT for power estimation is trained, the training time for each epoch is 10 minutes, and the inference time is about 0.9 seconds per batch.

%% file: appendix/appendix-B.tex
\section{CDFG of Hardware Designs}
\label{ap:cdfg}

\subsection{More CDFG Examples}
\begin{figure}[h]
    \centering
    \includegraphics[width=0.9\linewidth]{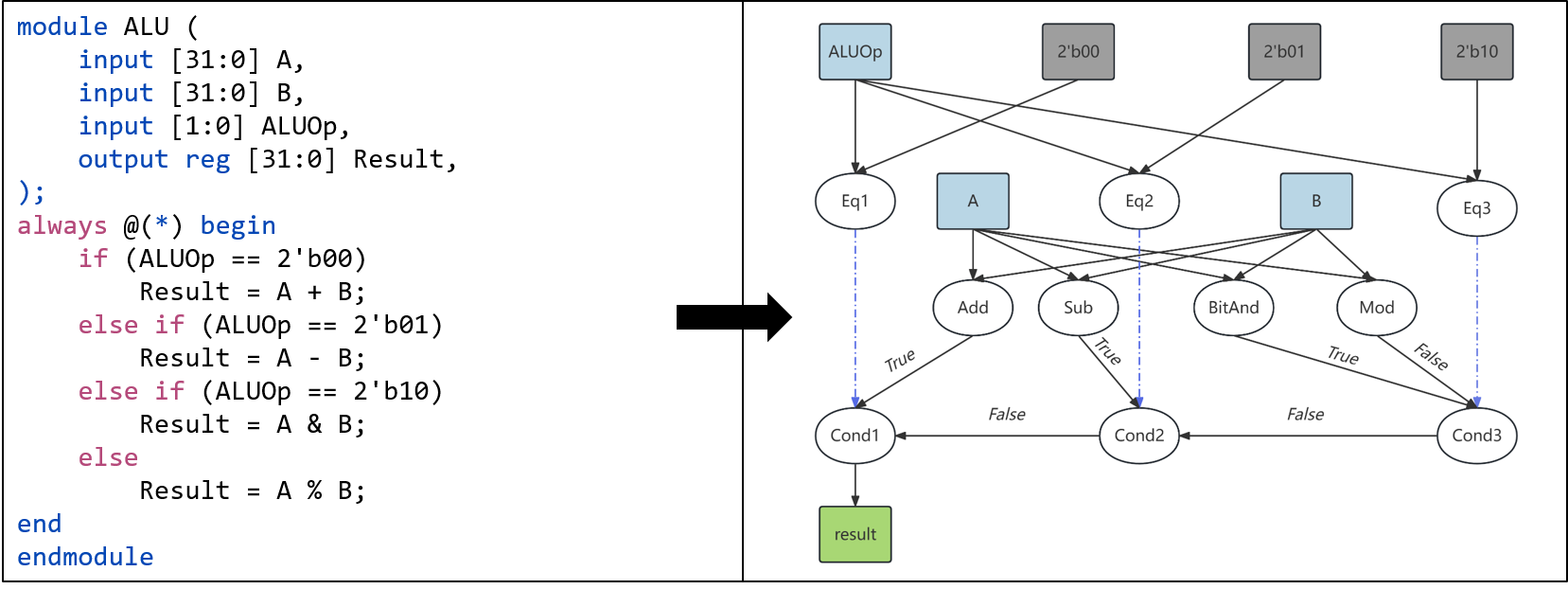}
    \vspace{-5pt}
    \caption{Arithmetic logic unit.}
    \vspace{-20pt}
    \label{fig:cdfg-ap1}
\end{figure}

\begin{figure}[h]
        \centering
        \includegraphics[width=0.9\linewidth]{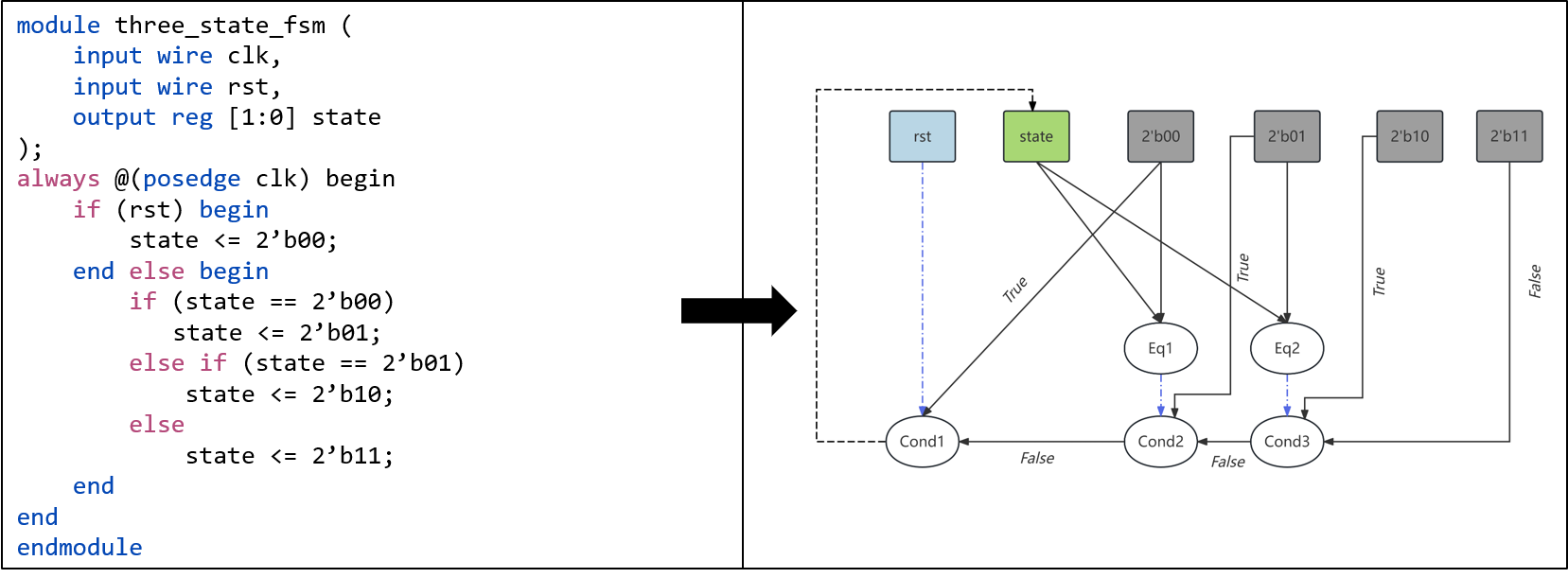}
        \vspace{-5pt}
        \caption{Three-state finite state machine.}
        \vspace{-20pt}
        \label{fig:cdfg-ap2}
\end{figure}

\begin{figure}[h]
        \centering
        \includegraphics[width=0.9\linewidth]{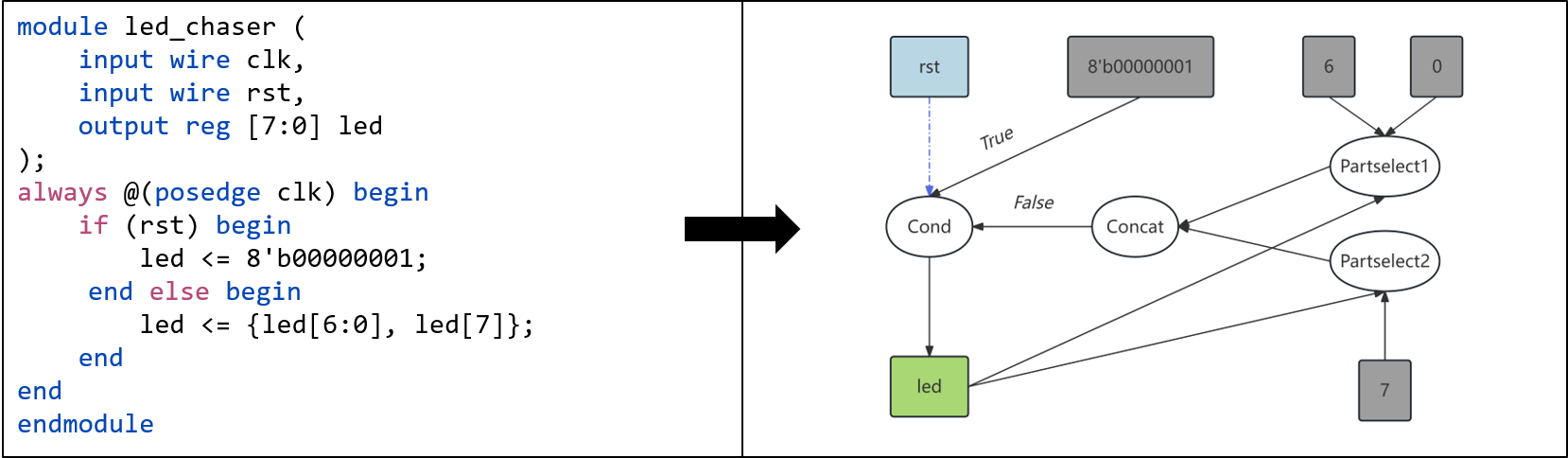}
        \vspace{-5pt}
        \caption{LED chaser.}
        \vspace{-20pt}
        \label{fig:cdfg-ap3}
\end{figure}

Figure~\ref{fig:cdfg-ap1}, ~\ref{fig:cdfg-ap2} and~\ref{fig:cdfg-ap3} illustrate more examples of the operator-level circuit control data flow graph.

\vspace{+100pt}

\FloatBarrier

\subsection{CDFG Node Sub-Types}
We describe the CDFG node sub-types in this subsection. Table~\ref{tab:cdfg-1} outlines the basic node types in CDFG. Table~\ref{tab:cdfg-unary} and ~\ref{tab:cdfg-binary} provides a detailed introduction of Verilog unary operators and binary operators, which are shortly mentioned in Table~\ref{tab:cdfg-1}. 

\vspace{10pt}

\renewcommand{\arraystretch}{1.8}
\begin{table}[ht]
  \centering
    \begin{tabular}{|c|c|c|p{6cm}|}
    \hline
    \textbf{Node-Type} & \textbf{In-Edge} & \textbf{Example} & \textbf{Description} \\ 
    \hline
    Unary Operator & Src  & \cmd{a = op1 b} & {Unary operator with single source operand.} \\
    \hline
    {Binary Operator} & {Src1, Src2}  & {\cmd{a = b op2 c}} & {Binary operator with two source operands.} \\
    \hline
    \multirow{4}*[+2ex]{PartSelect} & \multirow{4}*[+2ex]{Var, High, Low} & \multirow{4}*[+2ex]{\cmd{a = b[x:y]}}&{Select a part of a variable from the high index to low index.\newline{}\textit{Example: Assign bits from position x to y of b to a.}} \\
    \hline
    \multirow{3}*[+1ex]{Concat} & \multirow{3}*[+1ex]{Src1, Src2, ...}  & \multirow{3}*[+1ex]{\cmd{a = \{b,c,d\}}} & {Concatenate multiple variables together. \newline{}\textit{Example : Connect b, c, and d together in order and assign them to a.}} \\
    \hline
    \multirow{4}{*}[+2ex]{Condition} & \multirow{4}{*}[+2ex]{Sel, Data1, Data2}   & \multirow{4}{*}[+2ex]{\cmd{a = b?c:d}} & {Conditional selection between two data according to the select signal.\newline{}\textit{Example : If b is equal to True, assign c to a, otherwise assign d to a.}} \\
    \hline
    \multirow{3}* [+1ex]{Register}   &   \multirow{3}*[+1ex]{-}    & \multirow{3}*[+1ex]{\cmd{reg [7:0] a}} & { Register type variable in Verilog. \newline{}\textit{Example : An 8-bit register type variable named `a'. }} \\
    \hline
    \multirow{3}*[+1ex]{Wire}  &    \multirow{3}*[+1ex]{-}     & \multirow{3}*[+1ex]{\cmd{wire [7:0] b}} & {Wire type variable in Verilog.\newline{}\textit{Example : An 8-bit wire type variable named `b'.}} \\
    \hline
    \multirow{3}*[+1ex]{Constant} &    \multirow{3}*[+1ex]{-}     & \multirow{3}*[+1ex]{\cmd{32'd7}}  & {The constant in Verilog.\newline{}\textit{Example : A 32-bit constant representing decimal value 7.}} \\
    \hline
    \end{tabular}%
    \caption{Basic node type in circuit CDFG.}
    \label{tab:cdfg-1}
\end{table}%

\clearpage

\renewcommand{\arraystretch}{1.5}
\begin{table}[!h]
  \centering
    \begin{tabular}{|c|c|c|}
    \hline
    \textbf{~~~Operator-Type~~~} & \textbf{~~~~~~Example~~~~~~} & {\textbf{~~~~~~~~~~~Description~~~~~~~~~~~}} \\
    \hline
    LNot  & \cmd{!a} & {Logical Not} \\
    Not   & \cmd{\~{}a} & {Bitwise Not} \\
    URxor & \cmd{\^{}a} & {Xor on each bit} \\
    URand & \cmd{\&a} & {And on each bit} \\
    URor  & \cmd{|a} & {Or on each bit} \\
    URnand & \cmd{\~{}\&a} & {Nand on each bit} \\
    URnor  & \cmd{\~{}|a} & {Nor on each bit} \\
    \hline
    \end{tabular}%
    \caption{Detailed node types of unary operators in circuit CDFG.}
    \label{tab:cdfg-unary}
\end{table}%

\renewcommand{\arraystretch}{1.5}
\begin{table}[!h]
  \centering
    \begin{tabular}{|c|c|c|}
    \hline
    \textbf{~~~Operator-Type~~~} & \textbf{~~~~~~Example~~~~~~} & {\textbf{~~~~~~~~~~~Description~~~~~~~~~~~}} \\
    \hline
    Lt    & \cmd{a < b} & {Less than} \\
    Le    & \cmd{a <= b} & {Less than or equal to} \\
    Gt    & \cmd{a > b} & {Greater than} \\
    Ge    & \cmd{a >= b} & {Greater than or equal to} \\
    Add   & \cmd{b + c} & {-} \\
    Sub   & \cmd{b - c} &  {-}\\
    Mul   & \cmd{b * c} &  {-}\\
    Div   & \cmd{b / c} &  {-}\\
    Mod   & \cmd{b \% c} &  {-}\\
    ShiftLeft & \cmd{b << c} & {Logical left shift} \\
    ShiftRight & \cmd{b >> c} & {Logical right shift} \\
    AshiftLeft & \cmd{b <<< c} & {Arithmetic left shift} \\
    AshiftRight & \cmd{b >>> c} & {Arithmetic right shift} \\
    And   & \cmd{b \&\& c} & {-} \\
    Or    & \cmd{b || c} & {-} \\
    Eq    & \cmd{b == c} & {-} \\
    Neq   & \cmd{b != c} & {-} \\
    BitAnd & \cmd{b \& c} &  {Bitwise And}\\
    BitOr & \cmd{b | c} &  {Bitwise Or}\\
    BitXor & \cmd{b \^{} c} & {Bitwise Xor} \\
    \hline
    \end{tabular}%
    \caption{Detailed node types of binary operators in circuit CDFG.}
    \label{tab:cdfg-binary}
\end{table}%

\newpage


%% file: appendix/appendix-C.tex
\section{Circuit Dataset Details}
\label{ap:dataset}
This section provides a supplementary introduction to the DynamicRTL circuit dataset. We illustrate the scale of the circuit in our dataset. The scale of these circuit designs is comparable to prior circuit representation works~\cite{ deepgate_2, master_rtl, deepgate_3, deepseq}. We also showcase the diversity of our circuit dataset.

\begin{figure}[ht]
    \centering
    \includegraphics[width=0.45\linewidth]{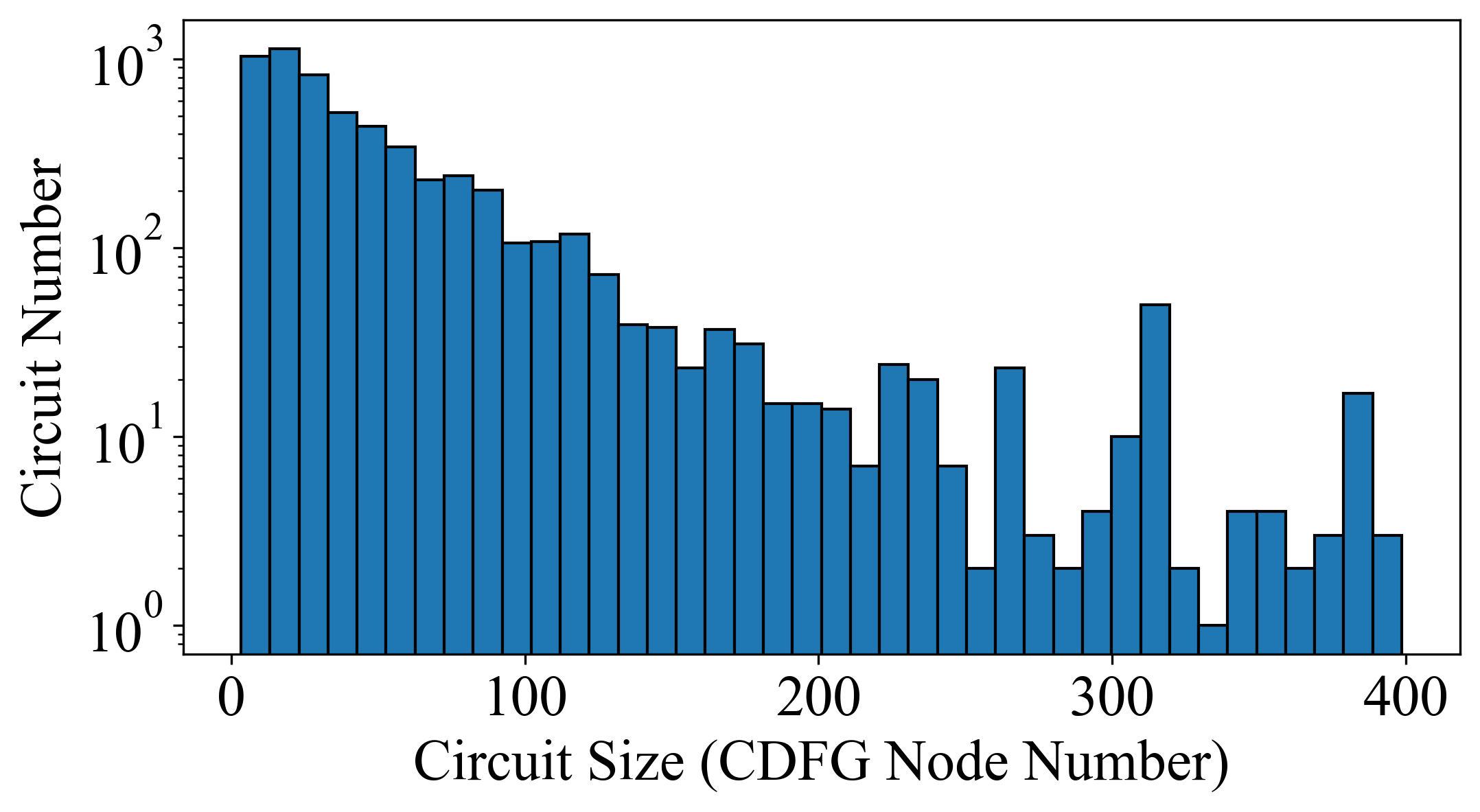}
    \caption{Histogram of circuit size (CDFG node number) in our dataset.}
    \label{fig:circuit_size}
    \vspace{+10pt}
\end{figure}

\textbf{Circuit Scale in DynamicRTL Dataset.}
The circuits in our dataset are stored as operator-level control data flow graphs (CDFGs). The designs range in size from 10 to over 500 CDFG nodes, with an average of 51 nodes per design.
In Figure~\ref{fig:circuit_size}, we illustrate the distribution of circuit sizes based on the number of nodes in the circuit control data flow graph.


Because CDFG is derived from circuit Register Transfer Level (RTL) descriptions and is a compact and high-level representation of circuit, its node count is much more reduced compared with lower-level circuit representations, such as single-bit operator graph~\cite{master_rtl} or netlist-level And-Inverter Graph (AIG)~\cite{deepgate_1,deepgate_2,deepgate_3}.
For example, an RTL-level CDFG with 50 nodes may expand to over 10,000 nodes at the netlist level after synthesis. 
Table~\ref{tab:circuit_scale_2} illustrates the actual scale of circuits in our dataset after synthesis. Table~\ref{tab:circuit_scale_2} (a) shows the code line number of original circuit designs in our dataset. Table~\ref{tab:circuit_scale_2} (b) shows the code line number after being synthesized to RTLIL by Yosys~\cite{yosys}. Table~\ref{tab:circuit_scale_2} (c) shows the node number after being synthesized to And Invert Graph (AIG) by Yosys. 

\input{tables/circuit_scale-exp3}
\textbf{Circuit Scale in Datasets of Previous Studies.}
DeepGate~\cite{deepgate_1} trains and tests on datasets with AIG nodes ranging from 36 to 3.2K. DeepGate2~\cite{deepgate_2} trains on the same dataset as DeepGate but tests its performance on larger designs with sizes from 13.2K to 47.3K AIG nodes. DeepSeq~\cite{deepseq} trains on circuits with several hundred AIG nodes and tests on large designs with sizes from 2.0K to 18.2K AIG nodes. HOGA~\cite{hoga} trains and tests on circuits with sizes ranging from 462 to 241K AIG nodes. FGNN~\cite{fgnn} trains on circuits with gate sizes from 138K to 143K and tests on circuits with gate sizes from 24.3K to 26.1K.
While DynamicRTL focuses on circuits at the RTL level CDFG, the circuits in our dataset have a comparable scale in both training and testing datasets compared to these previous works, with up to 150K nodes when translated to AIG.

\begin{figure}[ht]
    \centering
    \includegraphics[width=0.45\linewidth]{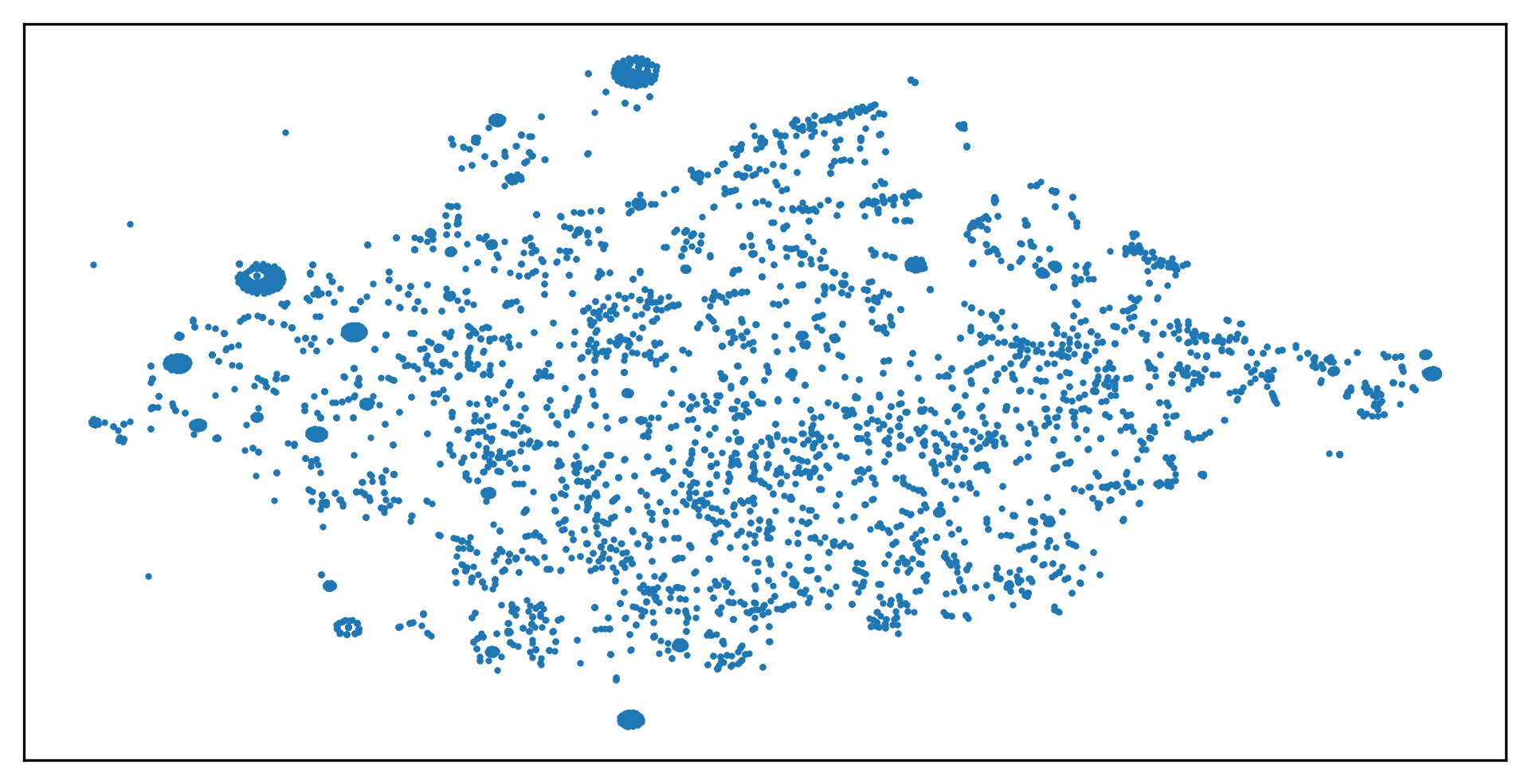}
    \caption{T-SNE visualization of circuit dataset.}
    \label{fig:tsne}
\end{figure}

\textbf{Circuit Diversity in DynamicRTL Dataset.}
To ensure the diversity of our circuit dataset and eliminate potential biases towards specific circuit types, we employed t-Distributed Stochastic Neighbor Embedding (t-SNE) for visualization. The circuit feature consists of the counts of each type of nodes present in the circuit. The resulting mapping, as illustrated in the Figure~\ref{fig:tsne}, shows that most circuits exhibit an average distribution, thereby demonstrating the diversity of our dataset.

%% file: tables/circuit_scale-exp3.tex
\begin{table}[h!]

\centering
\begin{subtable}[t]{0.23\textwidth}
\centering
\begin{tabular}{lc}
\toprule
\textbf{Range} & \textbf{Count} \\
\midrule
(0, 100]   & 3172 \\
(100, 200] & 1458 \\
(200, 500] & 1023 \\
(500, 1000] & 351 \\
1000+      & 320 \\
\bottomrule
\end{tabular}
\caption{Lines of Verilog}
\end{subtable}
\hspace{0.05\textwidth}
\begin{subtable}[t]{0.23\textwidth}
\centering
\begin{tabular}{lc}
\toprule
\textbf{Range} & \textbf{Count} \\
\midrule
(0, 100]   & 3048 \\
(100, 200] & 1517 \\
(200, 500] & 1002 \\
(500, 1000] & 389 \\
1000+      & 368 \\
\bottomrule
\end{tabular}
\caption{Lines of Verilog RTLIL}
\end{subtable}
\hspace{0.05\textwidth}
\begin{subtable}[t]{0.25\textwidth}
\centering
\begin{tabular}{lc}
\toprule
\textbf{Range} & \textbf{Count} \\
\midrule
(0, 200]      & 2469 \\
(200, 500]    & 1371 \\
(500, 1000]   & 932 \\
(1000, 2000]  & 653 \\
(2000, 5000]  & 450 \\
(5000, 10000] & 125 \\
(10000, 50000] & 142 \\
50000+        & 182 \\
\bottomrule
\end{tabular}
\caption{Nodes of AIG}
\end{subtable}

\caption{Circuit scale at Verilog, RTLIL and AIG level.}
\label{tab:circuit_scale_2}
\end{table}

%% file: sections/6-realted-work.tex
\section{Related Work}
\textbf{Learning Representation for Digital Circuits.}
Circuits can be represented as graphs, with operators as nodes and wires as edges, making GNNs useful for learning circuit representations~\cite{eda_gnn_survey}. 
Recent studies have focused on learning functionality-aware circuit representations.
FGNN~\citep{fgnn} uses contrastive learning to distinguish functionally equivalent circuits. 
DeepGate family~\cite{deepgate_1, deepgate_2, deepgate_3} transforms circuit into And-Inverter Graph (AIG) and uses node logic-1 probability and truth table similarity as training supervisions. However, these methods are limited to combinational circuits, falling short in sequential circuits.
Moreover, while these representations in netlist level are beneficial for tasks like logic synthesis and PPA prediction, they are less useful in the language front-end of circuit design, which requires RTL-level representations. MasterRTL~\cite{master_rtl} constructs a pre-synthesis PPA estimation framework, but its representation is at bit-level simple operator graph, not an original RTL format.

Our work focuses on learning dynamic representations of RTL designs with sequential behaviors. The most related study is Design2Vec~\citep{design2vec}, which predicts the branch coverage under specific test parameters. However, Design2Vec trains separate GNNs for each design, lacking a universal circuit representation. Moreover, its statement-level CDFG is code-oriented, where each node in graph represents a statement. While this approach effectively captures RTL code semantics, it cannot clearly reflects dynamic circuit data flow. 
DeepSeq~\citep{deepseq} also targets sequential circuits, which operates on netlist-level AIGs and predicts the logic-1 and state transition probabilities under random workloads. Without considering the different circuit behaviors under different input sequences, its representation significantly lacks dynamic information compared to ours. Meanwhile, our work learns circuits at RTL level, providing a more high-level graph representation compared to netlist-level representation methods.


\textbf{Learning Representation for Software Programs.} 
Recent studies have applied machine learning to software code representation~\citep{code_learning_survey}. Some studies focus on understanding dynamic behavior of programs, a concept known as \textit{learning to execute}. ~\citet{learn_to_execute} firstly uses RNN to read the program character-by-character and computes the program's output. ~\citet{ipagnn} transforms programs into statement-level control flow graphs, using a custom IPA-GNN for output prediction. ~\citet{learn_exe_code_fusion} focuses on assembly code, converting it into instruction-level control flow graphs and using GNN to predict the next execution branch and prefetch address.

Compared to hardware description code, software programs predominantly follow a serial execution logic, executing one statement at a time. In contrast, hardware code represents concurrent data flows between hardware components.
Furthermore, circuit RTL code is executed in each clock cycle, leading to dynamic behaviors (\ie state transition space) with much deeper depth. This makes learning dynamic hardware representation more challenging than software.


%% file: aaai2026.bib
@article{lcm_survey,
author = {Chen, Lei and Chen, Yiqi and Chu, Zhufei and Fang, Wenji and Ho, Tsung-Yi and Huang, Ru and Huang, Yu and Khan, Sadaf and Li, Min and Li, Xingquan and Li, Yu and Liang, Yun and Liu, Jinwei and Liu, Yi and Lin, Yibo and Luo, Guojie and others},
journal = {Science China Information Sciences},
month = {Oct.},
number = {10},
pages = {200402},
title = {{Large circuit models: opportunities and challenges}},
volume = {67},
year = {2024}
}

@inproceedings{eda_gnn_survey,
  title={Understanding graphs in {EDA}: From shallow to deep learning},
  author={Ma, Yuzhe and He, Zhuolun and Li, Wei and Zhang, Lu and Yu, Bei},
  booktitle={Proceedings of the International Symposium on Physical Design (ISPD)},
  pages={119--126},
  year={2020}
}

@inproceedings{master_rtl,
  title={{MasterRTL}: A Pre-Synthesis {PPA} Estimation Framework for Any {RTL} Design},
  author={Fang, Wenji and Lu, Yao and Liu, Shang and Zhang, Qijun and Xu, Ceyu and Wills, Lisa Wu and Zhang, Hongce and Xie, Zhiyao},
  booktitle={IEEE/ACM International Conference on Computer Aided Design (ICCAD)},
  pages={1--9},
  year={2023},
  _organization={IEEE}
}

@inproceedings{sengupta2022good,
  title={How good is your {Verilog} {RTL} code? {A} quick answer from machine learning},
  author={Sengupta, Prianka and Tyagi, Aakash and Chen, Yiran and Hu, Jiang},
  booktitle={Proceedings of the 41st IEEE/ACM International Conference on Computer-Aided Design (ICCAD)},
  pages={1--9},
  year={2022}
}

@inproceedings{lopera2021rtl,
  title={{RTL} delay prediction using neural networks},
  author={Lopera, Daniela S{\'a}nchez and Servadei, Lorenzo and Kasi, Vishwa Priyanka and Prebeck, Sebastian and Ecker, Wolfgang},
  booktitle={IEEE Nordic Circuits and Systems Conference (NorCAS)},
  pages={1--7},
  year={2021},
  _organization={IEEE}
}

@inproceedings{deepgate_1,
  title={{DeepGate}: Learning neural representations of logic gates},
  author={Li, Min and Khan, Sadaf and Shi, Zhengyuan and Wang, Naixing and Yu, Huang and Xu, Qiang},
  booktitle={Proceedings of the 59th ACM/IEEE Design Automation Conference (DAC)},
  pages={667--672},
  year={2022}
}

@inproceedings{deepgate_2,
  title={{DeepGate2}: Functionality-aware circuit representation learning},
  author={Shi, Zhengyuan and Pan, Hongyang and Khan, Sadaf and Li, Min and Liu, Yi and Huang, Junhua and Zhen, Hui-Ling and Yuan, Mingxuan and Chu, Zhufei and Xu, Qiang},
  booktitle={IEEE/ACM International Conference on Computer Aided Design (ICCAD)},
  pages={1--9},
  year={2023},
  _organization={IEEE}
}

@article{deepgate_3,
  title={{DeepGate3}: Towards Scalable Circuit Representation Learning},
  author={Shi, Zhengyuan and Zheng, Ziyang and Khan, Sadaf and Zhong, Jianyuan and Li, Min and Xu, Qiang},
  journal={arXiv preprint arXiv:2407.11095},
  year={2024}
}

@inproceedings{yosys,
  title={Yosys - A free {Verilog} synthesis suite},
  author={Wolf, Clifford and Glaser, Johann and Kepler, Johannes},
  booktitle={Proceedings of the 21st Austrian Workshop on Microelectronics (Austrochip)},
  volume={97},
  year={2013}
}

@inproceedings{stagira,
  title={Incremental {Verilog} Parser},
  author={Chen, Xiangli and Meng, Yuehua and Chen, Gang},
  booktitle={International Symposium of Electronics Design Automation (ISEDA)},
  pages={236--240},
  year={2023},
  _organization={IEEE}
}

@inproceedings{fgnn,
  title={Functionality matters in netlist representation learning},
  author={Wang, Ziyi and Bai, Chen and He, Zhuolun and Zhang, Guangliang and Xu, Qiang and Ho, Tsung-Yi and Yu, Bei and Huang, Yu},
  booktitle={Proceedings of the 59th ACM/IEEE Design Automation Conference (DAC)},
  pages={61--66},
  year={2022}
}

@article{design2vec,
  title={Learning semantic representations to verify hardware designs},
  author={Vasudevan, Shobha and Jiang, Wenjie Joe and Bieber, David and Singh, Rishabh and Ho, C Richard and Sutton, Charles and others},
  journal={Advances in Neural Information Processing Systems (NeurIPS)},
  volume={34},
  pages={23491--23504},
  year={2021}
}

@inproceedings{deepseq,
  title={{DeepSeq}: Deep Sequential Circuit Learning},
  author={Khan, Sadaf and Shi, Zhengyuan and Li, Min and Xu, Qiang},
  booktitle={Design, Automation \& Test in Europe Conference \& Exhibition (DATE)},
  pages={1--2},
  year={2024},
  _organization={IEEE}
}

@inproceedings{deepseq2,
  title={Deepseq2: Enhanced sequential circuit learning with disentangled representations},
  author={Khan, Sadaf and Shi, Zhengyuan and Zheng, Ziyang and Li, Min and Xu, Qiang},
  booktitle={Proceedings of the 30th Asia and South Pacific Design Automation Conference},
  pages={498--504},
  year={2025}
}

@article{code_learning_survey,
  title={A survey of machine learning for big code and naturalness},
  author={Allamanis, Miltiadis and Barr, Earl T and Devanbu, Premkumar and Sutton, Charles},
  journal={ACM Computing Surveys (CSUR)},
  volume={51},
  number={4},
  pages={1--37},
  year={2018},
  publisher={ACM New York, NY, USA}
}

@article{learn_to_execute,
  title={Learning to execute},
  author={Zaremba, Wojciech and Sutskever, Ilya},
  journal={arXiv preprint arXiv:1410.4615},
  year={2014}
}

@article{ipagnn,
  title={Learning to execute programs with instruction pointer attention graph neural networks},
  author={Bieber, David and Sutton, Charles and Larochelle, Hugo and Tarlow, Daniel},
  journal={Advances in Neural Information Processing Systems (NeurIPS)},
  volume={33},
  pages={8626--8637},
  year={2020}
}

@inproceedings{learn_exe_code_fusion,
author = {Shi, Zhan and Swersky, Kevin Jordan and Tarlow, Danny and Ranganathan, Parthasarathy and Hashemi, Milad},
booktitle = {International Conference on Learning Representations (ICLR)},
title = {Learning Execution through Neural Code Fusion},
year = {2020}
}

@inproceedings{mgverilog,
  title={{MG-Verilog}: Multi-grained Dataset Towards Enhanced {LLM}-assisted {Verilog} Generation},
  author={Zhang, Yongan and Yu, Zhongzhi and Fu, Yonggan and Wan, Cheng and Lin, Yingyan (Celine)},
  booktitle={The First IEEE International Workshop on LLM-Aided Design (LAD)}, 
  year={2024}
}

@inproceedings{hgt,
  title={Heterogeneous graph transformer},
  author={Hu, Ziniu and Dong, Yuxiao and Wang, Kuansan and Sun, Yizhou},
  booktitle={Proceedings of the Web Conference (WWW)},
  pages={2704--2710},
  year={2020}
}

@article{VeriGen,
author = {Thakur, Shailja and Ahmad, Baleegh and Pearce, Hammond and Tan, Benjamin and Dolan-Gavitt, Brendan and Karri, Ramesh and Garg, Siddharth},
title = {{VeriGen}: A Large Language Model for {Verilog} Code Generation},
year = {2024},
issue_date = {May 2024},
publisher = {Association for Computing Machinery},
address = {New York, NY, USA},
volume = {29},
number = {3},
_issn = {1084-4309},
_url = {https://doi.org/10.1145/3643681},
_doi = {10.1145/3643681},
journal = {ACM Trans. Des. Autom. Electron. Syst. (TODAES)},
month = apr,
articleno = {46},
numpages = {31},
keywords = {Transformers, verilog, GPT, large language models, EDA}
}

@article{gcn,
  title={Semi-supervised classification with graph convolutional networks},
  author={Kipf, Thomas N and Welling, Max},
  journal={arXiv preprint arXiv:1609.02907},
  year={2016}
}

@article{gat,
  title={Graph attention networks},
  author={Veli{\v{c}}kovi{\'c}, Petar and Cucurull, Guillem and Casanova, Arantxa and Romero, Adriana and Lio, Pietro and Bengio, Yoshua},
  journal={arXiv preprint arXiv:1710.10903},
  year={2017}
}

@inproceedings{gatv2,
  title={How Attentive are Graph Attention Networks?},
  author={Brody, Shaked and Alon, Uri and Yahav, Eran},
  booktitle={International Conference on Learning Representations},
  year={2021}
}

@article{gated_gcn,
  title={Residual gated graph convnets},
  author={Bresson, Xavier and Laurent, Thomas},
  journal={arXiv preprint arXiv:1711.07553},
  year={2017}
}

@article{number_repre,
  title={Neural execution engines: Learning to execute subroutines},
  author={Yan, Yujun and Swersky, Kevin and Koutra, Danai and Ranganathan, Parthasarathy and Hashemi, Milad},
  journal={Advances in Neural Information Processing Systems (NeurIPS)},
  volume={33},
  pages={17298--17308},
  year={2020}
}

@article{gt_recipe,
  title={Recipe for a general, powerful, scalable graph transformer},
  author={Ramp{\'a}{\v{s}}ek, Ladislav and Galkin, Michael and Dwivedi, Vijay Prakash and Luu, Anh Tuan and Wolf, Guy and Beaini, Dominique},
  journal={Advances in Neural Information Processing Systems (NeurIPS)},
  volume={35},
  pages={14501--14515},
  year={2022}
}

@inproceedings{nangate_45,
author = {Stine, James E and Castellanos, Ivan and Wood, Michael and Henson, Jeff and Love, Fred and Davis, W Rhett and Franzon, Paul D and Bucher, Michael and Basavarajaiah, Sunil and Oh, Julie and Jenkal, Ravi},
booktitle = {IEEE International Conference on Microelectronic Systems Education (MSE)},
_doi = {10.1109/MSE.2007.44},
_isbn = {0-7695-2849-X},
_month = {Jun.},
pages = {173--174},
_publisher = {IEEE},
title = {{FreePDK: An Open-Source Variation-Aware Design Kit}},
url = {https://si2.org/open-cell-library/},
year = {2007},
note = {Accessed: 2025-01-20},
}

@article{global_pos,
  title={A generalization of transformer networks to graphs},
  author={Dwivedi, Vijay Prakash and Bresson, Xavier},
  journal={arXiv:2012.09699 [cs.LG], AAAI 2021 Workshop on Deep Learning on Graphs: Methods and Applications (DLG-AAAI)},
  year={2021}
}

@article{relative_pos,
  title={Distance encoding: Design provably more powerful neural networks for graph representation learning},
  author={Li, Pan and Wang, Yanbang and Wang, Hongwei and Leskovec, Jure},
  journal={Advances in Neural Information Processing Systems (NeurIPS)},
  volume={33},
  pages={4465--4478},
  year={2020}
}

@inproceedings{hoga,
  title={Less is more: Hop-wise graph attention for scalable and generalizable learning on circuits},
  author={Deng, Chenhui and Yue, Zichao and Yu, Cunxi and Sarar, Gokce and Carey, Ryan and Jain, Rajeev and Zhang, Zhiru},
  booktitle={Proceedings of the 61st ACM/IEEE Design Automation Conference},
  pages={1--6},
  year={2024}
}
